\documentclass[runningheads]{llncs}

\usepackage[T1]{fontenc}
\usepackage[latin9]{inputenc}
\usepackage{array}
\usepackage{rotating}
\usepackage{booktabs}
\usepackage{multirow}
\usepackage{amsmath}
\usepackage{graphicx}
\usepackage{amssymb}
\usepackage{hyperref}
\usepackage{lineno}

\usepackage{xfrac,nicefrac}
\usepackage{microtype}
\usepackage{mathpazo}
 
\usepackage{color}
\usepackage[table]{xcolor}
\definecolor{mgy}{gray}{0.85}

\providecommand{\tabularnewline}{\\}

\usepackage[T1,small,euler-digits]{eulervm}

\graphicspath{{Figures/}} 
\begin{document}

\title{Biclustering Algorithms Based on Metaheuristics: \\A Review}

\author{Ad\'{a}n {Jos\'{e}-Garc\'{i}a}\inst{1}, Julie Jacques\inst{2}, Vincent Sobanski\inst{3}, Clarisse Dhaenens\inst{1}}

\authorrunning{A. {Jos\'{e}-Garc\'{i}a}, J. Jacques, V. Sobanski, C. Dhaenens}
\institute{Univ. Lille, CNRS, Centrale Lille, UMR 9189 CRIStAL, F-59000 Lille, France\\
\and
Lille Catholic University, Facult\'{e} de Gestion, Economie et Sciences, France
\and
Univ. Lille, Inserm, CHU Lille, Institut Universitaire de France (IUF), U1286 - INFINITE - Institute for Translational Research in Inflammation, F-59000 Lille, France}

\maketitle
%%%%%%%%%%%%%%%%%%%%%%%%%%%%%%%%%%%%%%%%%%%%%%%%%
%%%%%%%%%%%%%%%%%%%%%%%%%%%%%%%%%%%%%%%%%%%%%%%%%
\begin{abstract} Biclustering is an unsupervised machine learning technique that simultaneously clusters rows and columns in a data matrix. Biclustering has emerged as an important approach and plays an essential role in various applications such as bioinformatics, text mining, and pattern recognition. However, finding significant biclusters is an NP-hard problem that can be formulated as an optimization problem. Therefore, different metaheuristics have been applied to biclustering problems because of their exploratory capability of solving complex optimization problems in reasonable computation time. Although various surveys on biclustering have been proposed, there is a lack of a comprehensive survey on the biclustering problem using metaheuristics. This chapter will present a survey of metaheuristics approaches to address the biclustering problem. The review focuses on the underlying optimization methods and their main search components: representation, objective function, and variation operators. A specific discussion on single versus multi-objective approaches is presented. Finally, some emerging research directions are presented.

\keywords{Metaheuristics  \and Biclustering \and Clustering}
\end{abstract}

%%%%%%%%%%%%%%%%%%%%%%%%%%%%%%%%%%%%%%%%%%%%%%%%%
%\setcounter{tocdepth}{2} 
%\tableofcontents
%\clearpage

%\linenumbers
%%%%%%%%%%%%%%%%%%%%%%%%%%%%%%%%%%%%%%%%%%%%%%%%%
%%%%%%%%%%%%%%%%%%%%%%%%%%%%%%%%%%%%%%%%%%%%%%%%%
%%%%%%%%%%%%%%%%%%%%%%%%%%%%%%%%%%%%%%%%%%%%%%%%%
\section{Introduction}
\begin{figure}[b!]
\begin{centering}
\includegraphics[width=0.85\textwidth]{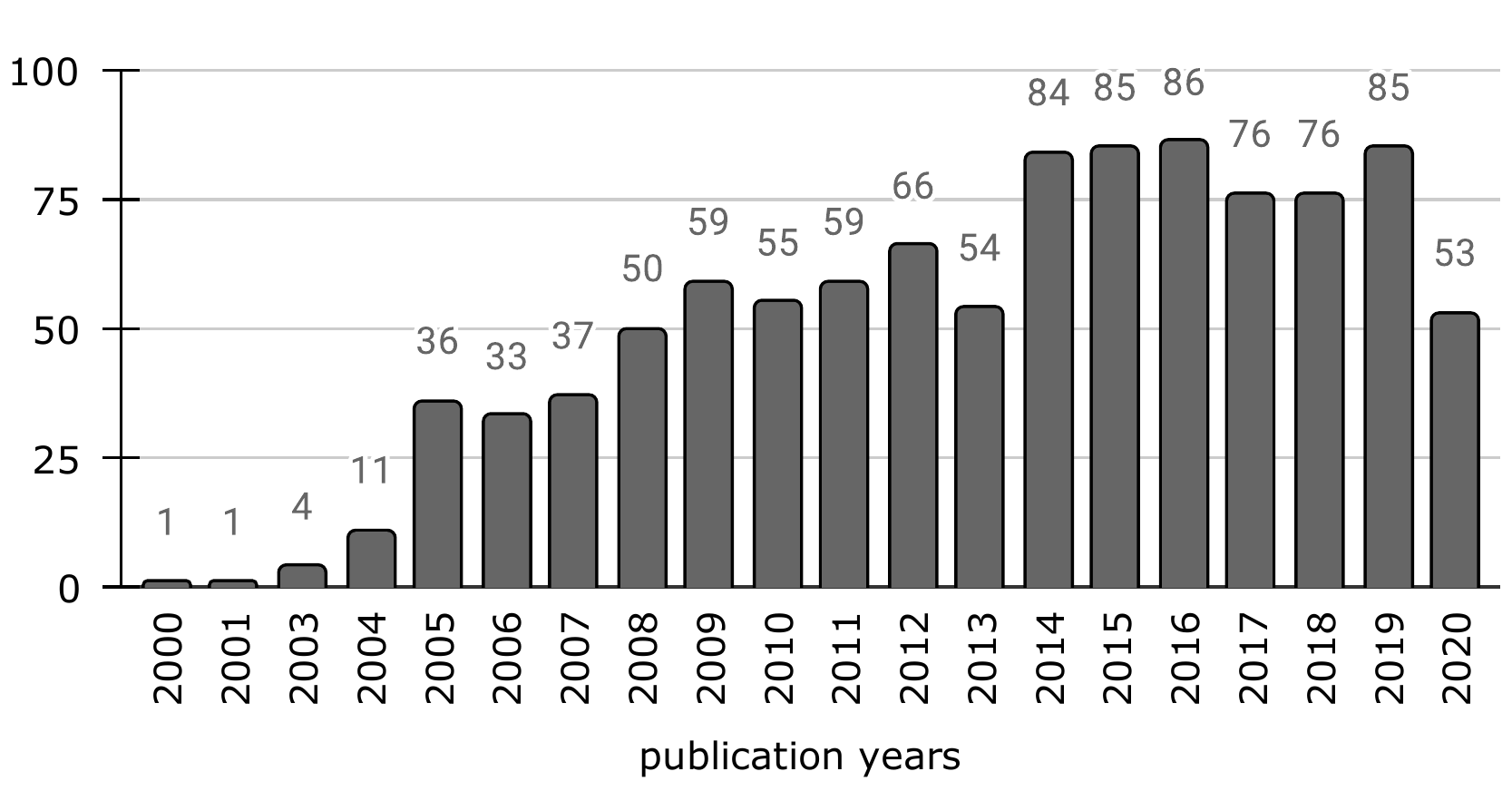}
\par\end{centering}
\caption{\label{fig:biclustering-publications}Number of studies on biclustering throughout the last 20 years. This bibliometric data was extracted from the Web of Science (WoS) database using the keyword \textquotedbl Biclustering.\textquotedbl{}}
\end{figure}

Biclustering is an unsupervised learning task to simultaneously cluster the rows and columns of a data matrix to obtain coherent and homogeneous biclusters (sub-matrices). For instance, in biological data, a subset of rows (or genes) is often correlated over a small subset of columns (or conditions). Biclustering\footnote{Biclustering is also referred to as co-clustering, subspace clustering, bi-dimensional or two-way clustering in the specialized literature.} is a helpful data mining technique that has gradually become a widely used technique in different applications such as bioinformatics, information retrieval, text mining, dimensionality reduction, recommendation systems, disease identification, and many more~\cite{Pontes2015,Xie2019}. In order to show the volume of studies currently been published in the field of biclustering, Figure~\ref{fig:biclustering-publications} presents a year-count of research publications about biclustering in the last 20 years. This plot shows an increasing trend in the number of publications since the first works in 2000.

%%%%%%%%%%%%%%%%%%%%%%%%%
%%%%%%%%%%%%%%%%%%%%%%%%%

The biclustering task has been proven to be an NP-complete\footnote{Nondeterministic Polynomial-time Complete: Any of a class of computational problems for which no efficient solution algorithm has been found.} problem~\cite{Cheng2000biclustering}, leading to propose optimization methods as heuristics to solve this combinatorial problem. Thus, developing an effective heuristic method and a suitable cost function are critical factors for discovering meaningful biclusters. In this regard, Pontes et al.~\cite{Pontes2015} analyzed a large number of approaches to biclustering and classified them depending on whether they use evaluation metrics within the search heuristic method. In particular, nature-inspired metaheuristics have gradually been applied successfully to biclustering problems because of their excellent exploratory capability~\cite{Pontes2015,Seridi2015,Nepomuceno2011}. These approaches model the behavior of natural phenomena, which exhibit an ability to learn or adapt to new situations to solve problems in complex and changing environments~\cite{Jose-Garcia2015}.

Several reviews on biclustering have been conducted, emphasizing different aspects and perspectives of the problem~\cite{madeira2004biclustering,Pontes2015,Xie2019}. However, despite the relevance of these review articles, to the best of our knowledge, no review paper on metaheuristics to biclustering has been published. Therefore, we present an up-to-date overview of metaheuristic-based biclustering approaches that have been reported in the last 15 years. This chapter contributes in the following three main aspects: (i)~it provides a review of important aspects of the biclustering task when addressed as an optimization problem such as encoding schemes, variation operators, and metric functions; (ii)~it presents an in-depth review on single-objective and multi-objective metaheuristics applied to biclustering; and (iii)~it gives some research opportunities and future tendencies in the biclustering field.

The outline of this chapter is as follows. Section~\ref{sec:basic-preliminaries} describes the basic terms and concepts related to biclustering analysis. Section~\ref{sec:metaheuristics-components} outlines the main components involved in metaheuristic-based biclustering algorithms. Section~\ref{sec:single-objective-biclustering} reviews single-objective biclustering algorithms in which a unique cost function is optimized. Section~\ref{sec:multi-objective-biclustering} presents multi-objective biclustering metaheuristics, which optimize distinct cost functions simultaneously. Section~\ref{sec:discussion} presents a discussion and future tendencies in biclustering. Finally, the conclusions are given in Section~\ref{sec:conclusion}.

%%%%%%%%%%%%%%%%%%%%%%%%%%%%%%%%%%%%%%%%%%%%%%%%%
%%%%%%%%%%%%%%%%%%%%%%%%%%%%%%%%%%%%%%%%%%%%%%%%%
%%%%%%%%%%%%%%%%%%%%%%%%%%%%%%%%%%%%%%%%%%%%%%%%%
\section{Basic preliminaries~\label{sec:basic-preliminaries}}

In this section we define the problem of biclustering, present classical approaches and expose how this problem may be modeled as a combinatorial optimization one.

%%%%%%%%%%%%%%%%%%%%%%%%%%%%%%%%%%%%%%%%%%%%%%%%%
\subsection{Biclustering definition~\label{subsec:biclustering-definition}}

Given a data matrix $\mathbf{X}\in\mathbb{R}^{N\cdot M}$ where $N$ denotes the number of patterns (rows) and $M$ denotes the number of attributes (columns). Let us define a set of patterns as $R$ and the set of attributes as $C$; therefore, the matrix $\mathbf{X}_{R,C}=\left(R,C\right)$ denotes the full dataset $\mathbf{X}$. Thus, a bicluster is a subset of rows that exhibit similar behavior across a subset of columns, which is denoted as $B_{I,J}=\left(I,J\right)$ such that $I\subseteq R$ and $J\subseteq C$.

Depending on the application and nature of data, several \emph{types of biclusters} have been described in the literature~\cite{madeira2004biclustering}. In the following, let $B_{I,J}=\left(I,J\right)$ be a bicluster in which $b_{ij}$ refers to the value of the $i$-th pattern under the $j$-th attribute. We can identify four major types of biclusters:

\begin{itemize}
\item Biclusters with constant values on all rows and columns: $b_{ij}=\pi$
\item Biclusters with constant values on rows or columns.
\begin{itemize}
\item Constant rows: $b_{ij}=\pi+\alpha_{i}$ or $b_{ij}=\pi\times\alpha_{i}$
\item Constant columns: $b_{ij}=\pi+\beta_{j}$ or $b_{ij}=\pi\times\beta_{j}$
\end{itemize}
\item Biclusters with coherent values on both rows and columns~\cite{aguilar2005shifting}.
\begin{itemize}
\item Shifting model (additive): $b_{ij}=\pi+\alpha_{i}+\alpha_{j}$
\item Scaling model (multiplicative): $b_{ij}=\pi\times\beta_{i}\times\beta_{j}$
\end{itemize}
\item Bicluster with coherent evolutions: A subset of patterns (rows) is up-regulated or down-regulated coherently across subsets of attributes (columns) irrespective of their actual values; that is, in the same directions but with varying magnitude. In these scenarios, coherent-evolution of biclusters are difficult to model using a mathematical equation.
\end{itemize}

In the previous bicluster definitions, $\pi$ represents any constant value for $B$, $\alpha_{i}\left(1\leq i\leq\left|I\right|\right)$ and $\alpha_{j}\left(1\leq j\leq\left|J\right|\right)$ refers to the constant values used in the additive models for each pattern and attribute; and $\beta_{i}\left(1\leq i\leq\left|I\right|\right)$ and $\beta_{j}\left(1\leq j\leq\left|J\right|\right)$ corresponds to the constant values used in the multiplicative models.

In most real-world problems, the cluster analysis involves the extraction of several biclusters, where the relations between the biclusters are defined by two criteria, \textit{exclusivity} and \textit{exhaustivity}. The exclusivity criterion indicates that an element must belong to a single bicluster, whereas the exhaustivity criterion specifies that every element must be part of one or more biclusters. Commonly, exclusivity refers to the covering of the input matrix, while exhaustivity is associated with overlapping among biclusters.

%%%%%%%%%%%%%%%%%%%%%%%%%
\begin{figure}[t]
\includegraphics[width=1\textwidth]{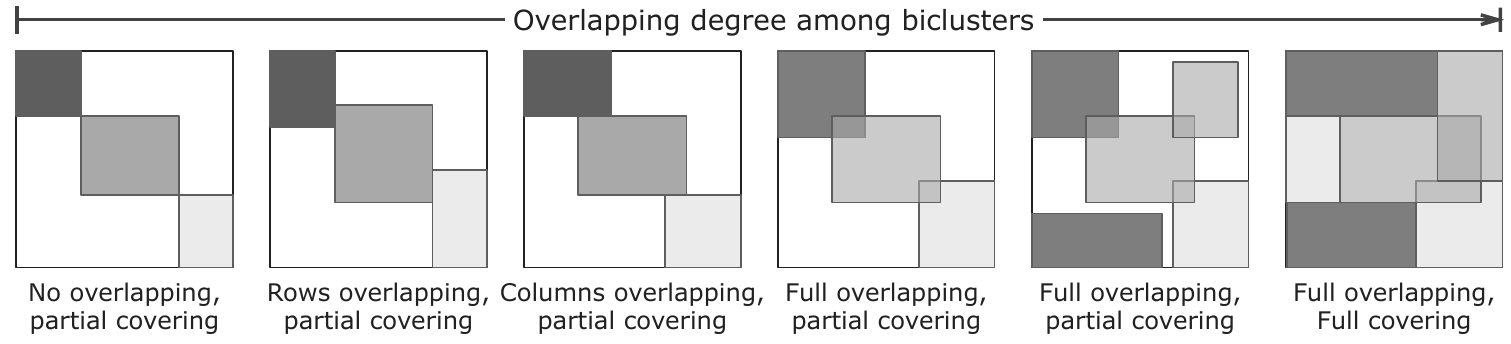}

\caption{\label{fig:bicluster-structures}Different bicluster structures based on exclusivity (matrix covering) and exhaustivity criteria (bicluster overlapping).}
\end{figure}
%%%%%%%%%%%%%%%%%%%%%%%%%

Considering these two criteria, several \emph{bicluster structures} can be obtained~\cite{madeira2004biclustering}. Figure~\ref{fig:bicluster-structures} illustrates some examples of bicluster structures, for instance, biclusters with: exclusive rows or exclusive columns (no overlapping and partial covering), exclusive rows and exhaustive columns (columns overlapping and partial covering), exhaustive rows and exhaustive columns (full overlapping and full covering).

Consequently, the selection of biclusters types and structures to be discovered will depend on both the problem being solved and the type of data involved. Therefore, the biclustering algorithm should be able to identify the desired biclusters.

%%%%%%%%%%%%%%%%%%%%%%%%%%%%%%%%%%%%%%%%%%%%%%%%%
\subsection{Classical biclustering approaches}

%%%%%%%%%%%%%%%%%%%%%%%%%
\begin{figure}[t!]
\includegraphics[width=1\textwidth]{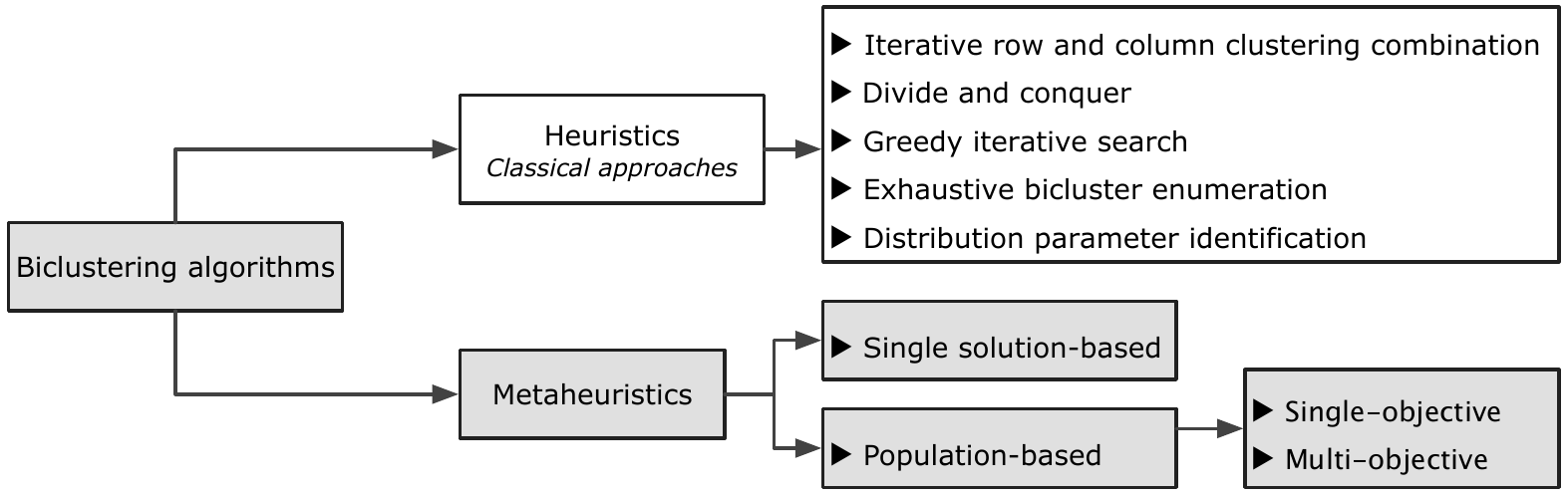}

\caption{\label{fig:biclustering-taxonomy}Biclustering algorithms taxonomy.}
\end{figure}
%%%%%%%%%%%%%%%%%%%%%%%%%

Biclustering has gained much interest since the seminal work of Cheng and Church on biclustering to analyze gene expression data~\cite{Cheng2000biclustering}. There are several biclustering algorithms in the literature, in particular, to deal with biological data, including detection of gene expression~\cite{Cheng2000biclustering,pontes2015biclustering}, protein interaction~\cite{ding2006biclustering}, and microarray data analysis~\cite{Kluger2003}.

To study such enriched literature, we categorized the biclustering algorithms into two main classes: \emph{heuristic-based} and \emph{metaheuristic-based} approaches. This categorization builds upon previous taxonomies proposed by Madeira and Oliveira~\cite{madeira2004biclustering} and by Seridi et al.~\cite{Seridi2015}. Figure~\ref{fig:biclustering-taxonomy} illustrates the employed taxonomy of biclustering algorithms, where the metaheuristic-based approaches are the primary focus of this review work and is detailed in the following sections. Regarding the heuristic-based algorithms, Madeira and Oliveira proposed to divide these approaches into five classes~\cite{madeira2004biclustering}:

\subsubsection*{Iterative row and column clustering combination}

To identify biclusters, classical clustering methods are applied to the rows and columns of the data matrix separately; then, the resulting clusters are combined using an iterative procedure to obtain biclusters of good quality. Among the algorithms adopting this approach are the Coupled Two-Way Clustering (CTWC)~\cite{Getz2000}, the Interrelated Two-Way Clustering (ITWC)~\cite{ChunTang2001}, and the Double Conjugated Clustering (DCC) algorithm~\cite{madeira2004biclustering}.

\subsubsection*{Divide-and-conquer}

These approaches split the problem into several smaller sub-problems of the same type, solving each one recursively. Then, the solutions of the sub-problems are combined to obtain a single solution to the original problem. Divide-and-conquer biclustering algorithms start with the entire data matrix as the initial bicluster. Then, this bicluster is split into several biclusters iteratively until satisfying a certain termination criterion. These approaches are quite fast, but good biclusters may be split before they can be identified. The main algorithms in this class are the Direct Clustering Algorithm (DCA)~\cite{Hartigan1972} and the Binary Inclusion-Maximal Biclustering Algorithm (Bimax)~\cite{Prelic2006}.

\subsubsection*{Greedy iterative search}

These methods create biclusters by adding or removing rows and columns using a quality criterion that maximizes the local gain. Therefore, although these approaches may make wrong decisions and miss good biclusters, they usually tend to be very fast. The most representative work in this category is the Cheng and Church's Algorithm (CCA)~\cite{Cheng2000biclustering}. Other approaches are the Order-Preserving Submatrix (OPSM)~\cite{Ben-Dor2003}, the QUalitative BIClustering (QUBIC)~\cite{Li2009}, and the Large Average Submatrices (LAS) algorithm~\cite{Shabalin2009}.

\subsubsection*{Exhaustive bicluster enumeration}

These algorithms consider that the best submatrices can only be identified by generating all the possible row and column combinations of the data matrix. Therefore, an exhaustive enumeration of all possible biclusters in the matrix is performed. These approaches find the best biclusters (if they exist), but they are suitable only for very small datasets. The main drawback is their high complexity, requiring restrictions on the size of the biclusters when performing the exhaustive enumeration. The main algorithms in this category are the Statistical-Algorithmic Method for Bicluster Analysis (SAMBA)~\cite{Tanay2002}, Bit-Pattern Biclustering Algorithm (BiBit)~\cite{Rodriguez-Baena2011}, and the Differentially Expressed Biclusters (DeBi)~\cite{Serin2011}.

\subsubsection*{Distribution parameter identification}

These approaches assume a given statistical model and try to identify the distribution parameters used to generate the data by optimizing a particular quality criterion. Some of the algorithms adopting this approach are Spectral Biclustering (BS)~\cite{Kluger2003}, the Bayesian BiClustering (BBC)~\cite{Gu2008}, and the Factor Analysis for Bicluster Acquisition (FABIA) algorithm~\cite{Hochreiter2010}.

%%%%%%%%%%%%%%%%%%%%%%%%%%%%%%%%%%%%%%%%%%%%%%%%%
\subsection{Biclustering as an optimization problem}

As mentioned previously, a bicluster $B\left(I,J\right)$ associated with a data matrix $\mathbf{X}\left(R,C\right)$ is a submatrix such that $I\subseteq R$ and $J\subseteq C$, where $R$ is a set of patterns (rows) and $C$ is a set of attributes (columns). The biclustering problem aims to extract biclusters of a maximal size that satisfy a coherence constraint. This task of extracting bicluster from a data matrix can be seen as a combinatorial optimization problem~\cite{dhaenens2016metaheuristics,dhaenens2019metaheuristics}.

Designing a bicluster is equivalent to jointly selecting a subset of rows and a subset of columns from an input data matrix $\mathbf{X}\in\mathbb{R}^{N\cdot M}$, where $N$ denotes the number of rows and $M$ denotes the number of columns. Let us assume that there is no restriction on the number of rows and columns and no constraints about the nature of the biclusters (i.e., bicluster type and structure). Then, by nature, the biclustering task is a combinatorial problem with a search space size of $\mathcal{O}\left(2^{N}*2^{M}\right)=\mathcal{O}\left(2^{N+M}\right)$.

However, identifying interesting biclusters is a complex task that requires defining some quality criteria to be optimized. Such criteria can measure the similarity within a bicluster, coherence, and dissimilarity (when a set of biclusters is searched). These biclustering quality criteria can be used as an objective function in a combinatorial optimization context, either alone or multiple in a multi-objective perspective. In the second scenario, several complementary quality criteria are optimized simultaneously.

In the general case, Cheng and Church showed that the problem of finding significant biclusters is NP-hard~\cite{Cheng2000biclustering}, giving rise to a large number of heuristic and metaheuristic approaches. These approaches do not guarantee the optimality of their solutions; however, their exploratory capabilities allow them to find suitable solutions in reasonable computation time. The following section is devoted to the review of metaheuristics designed to deal with biclustering problems.

%%%%%%%%%%%%%%%%%%%%%%%%%%%%%%%%%%%%%%%%%%%%%%%%%
%%%%%%%%%%%%%%%%%%%%%%%%%%%%%%%%%%%%%%%%%%%%%%%%%
%%%%%%%%%%%%%%%%%%%%%%%%%%%%%%%%%%%%%%%%%%%%%%%%%
\section{Main components of metaheuristics for biclustering~\label{sec:metaheuristics-components}}

%%%%%%%%%%%%%%%%%%%%%%%%%
\begin{figure}[t!]
\includegraphics[width=1\textwidth]{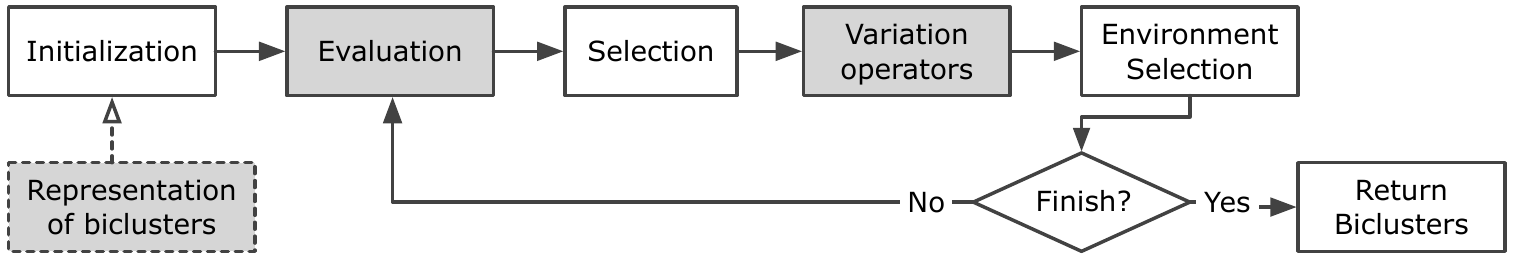}

\caption{\label{fig:metaheuristic-components}Main components of biclustering algorithms based on metaheuristics.}
\end{figure}
%%%%%%%%%%%%%%%%%%%%%%%%%

This section presents the main components involved when designing a metaheuristic to address the biclustering problem. First, a metaheuristic requires a suitable representation of biclusters, which is directly related to the objective function (i.e., bicluster quality measure) to be optimized. In general, the main steps in population-based metaheuristics are initializing the population, evaluating individuals, applying variation operators, and selecting the best biclustering solutions. This iterative process is repeated until a particular termination criterion is satisfied. Figure~\ref{fig:metaheuristic-components} illustrates these main components, where the main types of bicluster encodings, the variation operators to generate new biclusters, and different quality measures (evaluation step) are described below in this section. The other specific components, including the initialization procedures and selection strategies, are presented in detail when describing the single- or multi-objective biclustering algorithms in Sections~\ref{sec:single-objective-biclustering} and~\ref{sec:multi-objective-biclustering}.

%%%%%%%%%%%%%%%%%%%%%%%%%%%%%%%%%%%%%%%%%%%%%%%%%
\subsection{Bicluster encoding~\label{subsec:Bicluster-encoding}}

Metaheuristics require a representation or an encoding of potential solutions to the optimization problem. The encoding scheme is directly related to the objective function to be optimized and the recombination operators to generate new solutions. Therefore, the encoding schemes play a relevant role in the efficiency and effectiveness of any metaheuristic and constitute an essential step in its design.

In the literature, different encoding schemes are commonly used in metaheuristics to represent biclusters: \emph{binary encoding} and \emph{integer encoding}. These bicluster encodings are exemplified in Figure~\ref{fig:bicluster-types} by considering a $4\times3$ didactic data matrix and are separately described below.

\subsubsection*{Binary bicluster encoding (BBE)}

A bicluster $B(I,J)$ is represented as a binary vector of fixed-length size $\left(N+M\right)$: $x=\left\{ p_{1},\ldots,p_{N},s_{1},\ldots,s_{M}\right\} $, where the first $N$ positions are related to the number of patterns (rows) and the remaining $M$ positions to the number of attributes (columns) from the data matrix $\mathbf{X}$. If the $i$-th pattern or the $j$-th attribute in $\mathbf{X}$ belongs to the bicluster $B\left(I,J\right)$, then $p_{i}=1$ or $s_{j}=1$ for $1\leq i\leq N$ and $1\leq j\leq M$; otherwise, $p_{i}=0$ or $s_{j}=0$. Figure~\ref{fig:bicluster-types} illustrates the binary encoding for $N=4$ and $M=3$ when $x=\{1,0,1,0,0,1,1\}$.

\subsubsection*{Integer bicluster encoding (IBE)}

\begin{figure}[b!]
\begin{centering}
\includegraphics[width=1\textwidth]{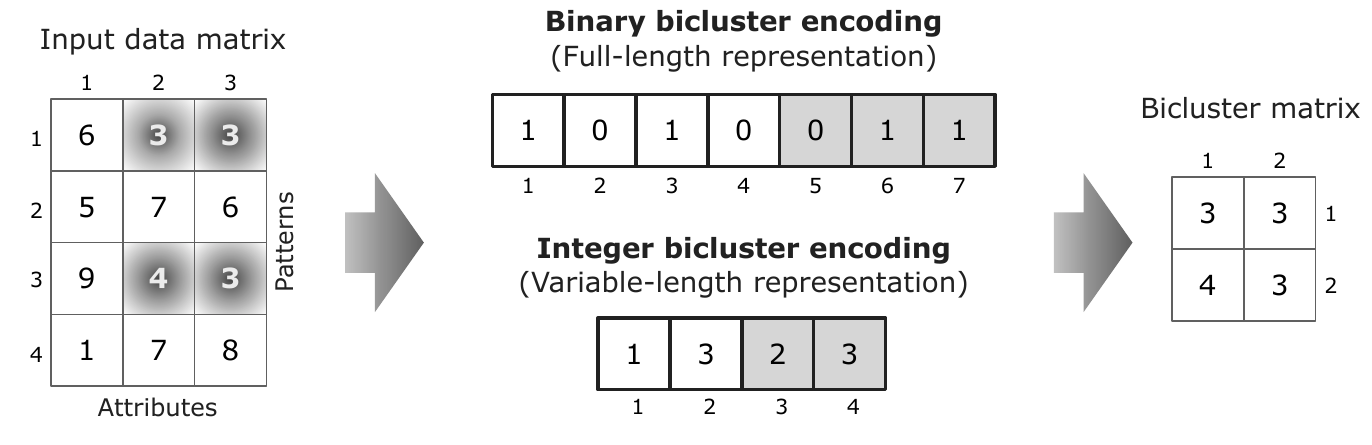}
\par\end{centering}
\caption{\label{fig:bicluster-types}Exemplification of two types of bicluster encodings in metaheuristic-based biclustering algorithms. The bicluster elements are shaded in the input data matrix, whereas the pattern and attribute indices are represented with white and gray boxes, respectively.}
\end{figure}

A bicluster $B(I,J)$ is represented as an integer vector of variable-length size $N_{p}+M_{s}$: $x=\left\{ p_{1},\ldots,p_{N_{p}},s_{1},\ldots,s_{M_{s}}\right\} $, where the first $N_{p}$ positions are ordered pattern indices, whereas the last $M_{s}$ positions correspond to ordered attribute indices from the data matrix $\mathbf{X}$. Each $i$-th pattern position takes an integer value in the set $\left\{ 1,\ldots,N\right\} $, and each $j$-th attribute position takes a value in $\left\{ 1,\ldots,M\right\} $, where $N$ and $M$ are the number of patterns and attributes in the data matrix, respectively~\cite{DeCastro2007}. Figure~\ref{fig:bicluster-types} illustrates the integer encoding for $N_{p}=2$ and $M_{s}=2$ when $x=\{1,3,2,3\}$.

Overall, the binary bicluster encoding is a practical representation, but it requires exploring all patterns and attributes of each bicluster. On the contrary, the integer encoding requires less computation time and memory space as it depends on the number of ordered patterns and attributes for a particular bicluster. Therefore, the integer bicluster encoding is more efficient in terms of time and space; however, it is more impractical when dealing with variable-length solutions in population-based metaheuristics.

%%%%%%%%%%%%%%%%%%%%%%%%%
%%%%%%%%%%%%%%%%%%%%%%%%%
%%%%%%%%%%%%%%%%%%%%%%%%%%%%%%%%%%%%%%%%%%%%%%%%%
\subsection{Variation operators}

In metaheuristic algorithms, variation operators play an important role in providing a good compromise between exploration and diversification to search solutions of high quality. In the following, we describe some crossover and mutation operators that are commonly used in biclustering algorithms.

\subsection*{Crossover operators}

The crossover is a basic component in metaheuristic biclustering methods, and in general, in evolutionary genetic algorithms. It tries to combine pair of individuals to produce new offsprings that inherits parent's features.

Let us consider two parent individuals $P_{1}=\left\{ r_{1},\ldots,r_{n},c_{1},\ldots,c_{m}\right\} $ and $P_{2}=\left\{ r_{1}^{\prime},\ldots,r_{l}^{\prime},c_{1}^{\prime},\ldots,c_{k}^{\prime}\right\} $, where $r_{n}\leq r_{l}^{\prime}$.

\subsubsection*{Single crossover operator~\cite{Seridi2011,Seridi2015}}

The crossover is performed in each part of the individual (rows part and columns part). The crossover point $\lambda_{1}$ in $P_{1}$ is generated randomly in the range $r_1<\lambda_{i}\leq r_{n}$. The random point in $P_{2}$, $\lambda_{2}=r_{j}^{\prime}$ where $r_{j}^{\prime}\geq\lambda_{1}$ and $r_{j-1}^{\prime}\leq\lambda_{1}$. The crossover in the columns part is performed similarly to the rows part.

\subsubsection*{Bicluster crossover operator~\cite{Maatouk2019}}

The crossover is based on the following four steps: (i)~creation of the merge bicluster $B_{\textrm{merge}}$ from two parent biclusters (individuals); (ii)~discretization of the merge bicluster, $B_{\textrm{discrete}}$; (iii)~construction of the variation matrix, \textbf{$M_{\textrm{var}}$}; (iv)~extraction of child biclusters.

\subsection*{Mutation operators}

Mutation operators aim to modify an individual, either randomly or using a specific strategy.

\subsubsection*{Random mutation operator~\cite{Maatouk2019}}

Let us consider an individual in the population $B=\left\{ r_{1},\ldots,r_{n},c_{1},\ldots,c_{m}\right\}$, two mutation points $r_{i}$ and $c_{j}$ are generated corresponding to the rows and columns parts, respectively, such that $r_1<r_{i}\leq r_{n}$ and $c_1<c_{j}\leq c_{m}$. Then, a random row $r_{i}^{\prime}$ and column $c_{j}^{\prime}$ values are generated to replace the chosen positions $r_{i}$ and $c_{j}$, respectively.

\subsubsection*{CC-based mutation operator~\cite{Seridi2011,Seridi2015}}

This mutation strategy is based on the CC algorithm~\cite{Cheng2000biclustering}, which aims to generate $k$ biclusters. Thus, given an individual, its irrelevant rows or columns having mean squared residue (MSR) values above (or below) a certain threshold are eliminated (or added) using the following conditions\footnote{A ``node'' refers to a row or a column.}: (i)~multiple node deletion, (ii)~single node deletion, and (iii)~multiple node addition.

%%%%%%%%%%%%%%%%%%%%%%%%%%%%%%%%%%%%%%%%%%%%%%%%%
\subsection{Objective functions~\label{subsec:objective-functions}}

Because of the complexity of the biclustering problem, several evaluation functions have been proposed to measure the quality of biclusters. This section presents some of the most well-known evaluation measures that have been used as objective functions in different metaheuristics.

For convenience, consider the following common notation for interpretation of the different objective functions. Given a bicluster $B(I,J)$ with $\left|I\right|$ patterns (rows), $\left|J\right|$ attributes (columns), the average $b_{iJ}$ of the $i$-th row, the average $b_{Ij}$ of $j$-th column, and the average value $b_{IJ}$ of the bicluster $B(I,J)$ are represented respectively as follows:

\begin{equation}
\begin{array}{ccccc}
b_{iJ}=\frac{1}{\left|J\right|}\sum_{j\in J}b_{ij}, &  & b_{Ij}=\frac{1}{\left|I\right|}\sum_{i\in I}b_{ij}, &  & b_{IJ}=\frac{1}{\left|I\right|\cdot\left|J\right|}\sum_{i\in I,j\in J}b_{ij}\,.\end{array}\label{eq:fit-bic}
\end{equation}

Additionally, the following notation is used, an abbreviation of the bicluster evaluation measure followed by an arrow to denote if the function is maximized ($\uparrow$) or minimized ($\downarrow$), indicating the bicluster solution's quality.

%%%%%%%%%%%%%%%%%%%%%%%%%
\subsubsection*{Bicluster Size (BSize)}

This function represents the size of a bicluster, where $\alpha$ is a constant representing a preference towards the maximization of the number of rows or columns:

\begin{equation}
\textrm{BSize}(B)_{\uparrow}=\alpha\frac{\left|I\right|}{\left|I'\right|}+\left(1-\alpha\right)\frac{\left|J\right|}{\left|J'\right|}\,.\label{eq:fit-bsize}
\end{equation}

Sometimes this measure is referred as the \emph{Volume} of a bicluster, i.e., as the number of elements $b_{ij}$ in $B$, $\left|I\right|\cdot\left|J\right|$.

%%%%%%%%%%%%%%%%%%%%%%%%%
\subsubsection*{Bicluster Variance (VAR)}

Hartigan~\cite{Hartigan1972} proposed the variance measure to identify biclusters with constant values:

\begin{equation}
\textrm{VAR}(B)_{\uparrow}=\sum_{i=1}^{\left|I\right|}\sum_{j=1}^{\left|J\right|}\left(b_{ij}-b_{IJ}\right)^{2}\,.\label{eq:fit-var}
\end{equation}

Existing biclustering approaches deal with biclustering variance in different ways. For instance, Pontes et al.~\cite{Pontes2013} used the average variance to avoid obtaining trivial biclusters. In practice, the row variance (rVAR) is used to avoid trivial or constant-value biclusters (i.e., biclusters with significant rows variances). The rVAR measure is defined as:
\begin{equation}
\textrm{rVAR}(B)_{\uparrow}=\frac{1}{\left|I\right|\cdot\left|J\right|}\sum_{i=1}^{\left|I\right|}\sum_{j=1}^{\left|J\right|}\left(b_{ij}-b_{iJ}\right)^{2}\,.\label{eq:fit-rvar}
\end{equation}

%%%%%%%%%%%%%%%%%%%%%%%%%
\subsubsection*{Mean Squared Residence (MSR)}

Cheng and Chung~\cite{Cheng2000biclustering} proposed the MSR to measure the correlation (or coherence) of a bicluster. It is defined as follows:

\begin{equation}
\textrm{MSR}(B)_{\downarrow}=\frac{1}{\left|I\right|\cdot\left|J\right|}\sum_{i=1}^{\left|I\right|}\sum_{j=1}^{\left|J\right|}\left(b_{ij}-b_{iJ}-b_{Ij}+b_{IJ}\right)^{2}\,.\label{eq:fit-mse}
\end{equation}
The lower the MSR, the stronger the coherence exhibited by the bicluster, and the better its quality. If a bicluster has an MSR value lower than a given threshold $\delta$, then it is called a $\delta$-bicluster.

%%%%%%%%%%%%%%%%%%%%%%%%%
\subsubsection*{Scaling Mean Squared Residence (SMSR)}

Mukhopadhyay et al.~\cite{Mukhopadhyay2009} developed this measure to recognize scaling patterns in biclusters. The SMSR measure is defined as:

\begin{equation}
\textrm{SMSR}(B)_{\downarrow}=\frac{1}{\left|I\right|\cdot\left|J\right|}\sum_{i=1}^{\left|I\right|}\sum_{j=1}^{\left|J\right|}\frac{\left(b_{ij}-b_{iJ}-b_{Ij}+b_{IJ}\right)^{2}}{b_{iJ}^{2}\cdot b_{Ij}^{2}}\,.\label{eq:fit-smse}
\end{equation}

%%%%%%%%%%%%%%%%%%%%%%%%%
\subsubsection*{Average Correlation Function (ACF)}

Nepomuceno et al.~\cite{Nepomuceno2010} proposed the ACF to evaluate the correlation between patterns in a bicluster. It is defined as:

\begin{equation}
\textrm{ACF}(B)_{\uparrow}=\frac{2}{\left|I\right|\left(\left|I\right|-1\right)}\sum_{i=1}^{\left|I\right|}\sum_{j=i+1}^{\left|I\right|}\left|\frac{cov\left(p_{i},p_{j}\right)}{\sigma_{p_{i}}\sigma_{p_{j}}}\right|\,,\label{eq:fit-acf}
\end{equation}
where $cov\left(p_{i},p_{j}\right)=\nicefrac{1}{\left|J\right|}\sum_{k=1}^{\left|J\right|}\left(b_{ik}-b_{iJ}\right)\left(b_{jk}-b_{jJ}\right)$ represents the covariance of the rows corresponding to patterns $p_{i}$ and $p_{j}$, and $\sigma_{p_{i}}$ (respectively, $\sigma_{p_{j}}$) represents the standard deviations of the rows corresponding to patterns $p_{i}$ and $p_{j}$. The ACF measure generates values in the range $\left[-1,1\right]$, where values close to the unity represents that the patterns in $B$ are highly correlated.

%%%%%%%%%%%%%%%%%%%%%%%%%
\subsubsection*{Average Correlation Value (ACV)}

Teng and Chan~\cite{Teng2008} proposed this measure to evaluate the correlation homogeneity of a bicluster. The ACV is defined as:

\begin{equation}
\textrm{ACV}(B)_{\uparrow}=\max\left\{ \frac{1}{\left|I\right|\left(\left|I\right|-1\right)}\sum_{i,j\in I,i\neq j}r_{ij},\frac{1}{\left|J\right|\left(\left|J\right|-1\right)}\sum_{i,j\in J,i\neq j}r'_{ij}\right\} \,,\label{eq:fit-acv}
\end{equation}
where $r_{ij}$ or $r'_{ij}$ is a Pearson correlation coefficient between the $i$-th row and $j$-th column. ACV generates values in the interval $\left[0,1\right]$, where a value closer to 1 indicates that the rows or columns in the bicluster are highly co-expressed, whereas a low ACV values means the opposite.

%%%%%%%%%%%%%%%%%%%%%%%%%
\subsubsection*{Virtual Error (VE)}

Divina et al.~\cite{Divina2012} proposed the VE function to identify shifting or scaling patterns in biclusters. It is defined as follows:

\begin{equation}
\textrm{VE}(B)_{\downarrow}=\frac{1}{\left|I\right|\cdot\left|J\right|}\sum_{i=1}^{\left|I\right|}\sum_{j=i}^{\left|J\right|}\left|\hat{b}_{ij}-\hat{\rho}_{i}\right|\,,\label{eq:fit-ve}
\end{equation}
where $\rho_{i}=\nicefrac{1}{j}\sum_{j=1}^{\left|J\right|}b_{ij}$ represents a \emph{virtual pattern} from $B$, and $\hat{b}_{ij}$ (respectively, $\hat{\rho}_{i}$) is the \emph{standardized value} of the element $b_{ij}$. The standardized values of a vector $V=\left\{ v_{1},\ldots,v_{n}\right\} $, denoted as $\hat{V}$, is the set $\hat{V}=\left\{ \hat{v}_{1},\ldots,\hat{v}_{n}\right\} $ with $\hat{v}_{k}=\left(v_{k}-\mu_{V}\right)\left/\sigma_{V}\right.$ for $1\leq k\leq n$, where $\mu_{V}$ and $\sigma_{V}$ are the mean and standard deviation of $V$. A small value of VE represents a high similarity among the patterns in the biclusters.

%%%%%%%%%%%%%%%%%%%%%%%%%
\subsubsection*{Coefficient of Variation Function (CVF)}

Maatouk et al.~\cite{Maatouk2014} presented the CVF to characterize the variability of a bicluster. The function is defined as follows:

\begin{equation}
CVF(B)_{\uparrow}=\frac{\sigma_{B}}{b_{IJ}}\,,\label{eq:fit-cvf}
\end{equation} 
where $\sigma_{B}$ represents the standard deviation of the bicluster and $b_{IJ}$ denotes the average of all the values in the bicluster $B$. A high value of CVF indicates that the bicluster presents a high level of dispersion.

%%%%%%%%%%%%%%%%%%%%%%%%%%%%%%%%%%%%%%%%%%%%%%%%%
%%%%%%%%%%%%%%%%%%%%%%%%%%%%%%%%%%%%%%%%%%%%%%%%%
%%%%%%%%%%%%%%%%%%%%%%%%%%%%%%%%%%%%%%%%%%%%%%%%%
\section{Single-objective biclustering~\label{sec:single-objective-biclustering}}

This section presents a review of single-objective metaheuristics for the biclustering problem. These approaches are mostly population-based metaheuristics that iteratively attempt to improve a population of biclustering solutions. First, the population is usually initialized randomly. Then, a new population of potential solutions is generated, which could be integrated into the current one by using some selection criteria. Finally, the search process stops when a given condition is satisfied (see Figure~\ref{fig:metaheuristic-components}).

Table~\ref{tab:single-objective-algs} summarizes relevant details of 23 single-objective biclustering algorithms based on metaheuristics. These biclustering algorithms, including hybrid biclustering metaheuristic approaches, are described in more detail below in this section.

%%%%%%%%%%%%%%%%%%%%%%%%%
\begin{table}[t!]
\small %\scriptsize %\small
\def\arraystretch{1.2}
\setlength{\tabcolsep}{0.01cm}

\caption{\label{tab:single-objective-algs}List of single-objective biclustering algorithms based on metaheuristics. \textquotedblleft Metaheuristic\textquotedblright{} indicates the type of nature-inspired metaheuristic, which can be simulated annealing (SA), genetic algorithm (GA), particle swarm optimization (PSO), scatter search (SS), and cuckoo search (CS). \textquotedblleft Objective Function\textquotedblright{} refers to the bicluster evaluation metric as defined in Section~\ref{subsec:objective-functions} (when several are indicated a linear combination of them is used); and \textquotedblleft Encoding\textquotedblright{} indicates the type of bicluster encoding binary (BBE) or integer (IBE) as described in Section~\ref{subsec:Bicluster-encoding}.}

\begin{tabular*}{1\textwidth}{@{\extracolsep{\fill}}rllccr}
\toprule 
\multirow{1}{*}{\textbf{Year}} & \textbf{Algorithm} & \textbf{Metaheuristic} & \textbf{Objective Function} & \textbf{Encoding} & \textbf{Ref.}\tabularnewline
\midrule
2004 & \texttt{HEA} & GA & MSR & BBE & \cite{Bleuler2004}\tabularnewline
2006 & \texttt{SAB} & SA & MSR & Other & \cite{Bryan2006}\tabularnewline
2006 & \texttt{SEBI} & GA & MSR, BSize, Var & BBE & \cite{Divina2006}\tabularnewline
2009 & \texttt{BiHEA} & GA & MSR & BBE & \cite{Gallo2009}\tabularnewline
2009 & \texttt{SS\&GA} & Hybrid: GA + SS & MSR & BBE & \cite{Nepomuceno2009}\tabularnewline
2010 & \texttt{HEAB} & Hybrid: GA + SS & ACF & BBE & \cite{Nepomuceno2010}\tabularnewline
2011 & \texttt{SSB} & SS & ACF & BBE & \cite{Nepomuceno2011}\tabularnewline
2011 & \texttt{BPSO} & PSO & ACV & BBE & \cite{Rathipriya2011}\tabularnewline
2012 & \texttt{CBEB} & GA & MSR & BBE & \cite{QinghuaHuang2012}\tabularnewline
2012 & \texttt{EvoBic} & GA & BSize, MSR, ACF & IBE & \cite{Ayadi2012}\tabularnewline
2012 & \texttt{PSO-SA-BIC} & Hybrid: PSO + SA & ACV & BBE & \cite{Thangavel2012}\tabularnewline
2013 & \texttt{Evo-Bexpa} & GA & VE, Vol, Overlap, Var & BBE & \cite{Pontes2013}\tabularnewline
2014 & \texttt{EBACross} & GA & BSize, MSR, ACF, CVF & BBE & \cite{Maatouk2014}\tabularnewline
2014 & \texttt{TriGen} & GA & MSR$_{\textrm{time}}$, LSL$_{\textrm{time}}$ & IBE & \cite{Gutierrez-Aviles2014}\tabularnewline
2015 & \texttt{COCSB} & CS & MSR & BBE & \cite{Lu2015}\tabularnewline
2015 & \texttt{SSB-Bio} & SS & BSize, ACF, ACF$_{\textrm{bio}}$ & BBE & \cite{Nepomuceno2015}\tabularnewline
2015 & \texttt{BISS} & SS & ACF & BBE & \cite{Nepomuceno2015a}\tabularnewline
2018 & \texttt{BISS-go} & SS & ACF$_{\textrm{go}}$ & BBE & \cite{Nepomuceno2018}\tabularnewline
2018 & \texttt{GACSB} & Hybrid: GA + CS & MRS, VE, ACV & BBE & \cite{Yin2018}\tabularnewline
2018 & \texttt{EBA} & GA & BSize, MSR, ACF, CVF & BBE & \cite{Maatouk2019}\tabularnewline
2019 & \texttt{BP-EBA} & GA & MSR, SMSR, BSize & BBE & \cite{Huang2019}\tabularnewline
2021 & \texttt{HPSO-TriC} & Hybrid: PSO + SA & ACF$_{\textrm{time}}$ & BBE & \cite{Narmadha2021}\tabularnewline
2021 & \texttt{ELSA} & GA & ACF$_{\textrm{stat}}$, ACF$_{\textrm{bio}}$ & BBE & \cite{Maatouk2021}\tabularnewline
\bottomrule
\end{tabular*}
$\left(*\right)$ The approaches that list more than one objective function indicate that the algorithms combine the information from these metrics into a single objective function.

%$\left(-\right)$ The \texttt{SAB} algorithm does not require an encoding strategy.
\end{table}
%%%%%%%%%%%%%%%%%%%%%%%%%

%%%%%%%%%%%%%%%%%%%%%%%%%%%%%%%%%%%%%%%%%%%%%%%%%
\subsection{Simulated annealing}

\emph{Simulated annealing} (SA) is a probabilistic method proposed by Kirkpatrick et~al.~\cite{Kirkpatrick1983} to find the global minimum of a cost function. SA emulates the physical process whereby a melted solid material (initial state) is gradually cooled until the minimum energy state is reached, that is when the material structure is ``frozen''. SA is a single-solution-based metaheuristic that improves a single point solution, evaluated by a single criterion function, and could be viewed as search trajectories through the search space~\cite{Jose-Garcia2015}.

Simulated annealing has been used to address the biclustering problem. Brayan et al. \cite{Bryan2006} proposed an SA-based biclustering approach called \textbf{SAB}. In SAB, each solution's fitness function is computed using the MSR criterion, and the algorithm is run $k$ times to obtain $k$ biclusters. In order to avoid overlap among biclusters, in the SAB algorithm, the discovered biclusters are masked in the original data. This strategy is similar to the Cheng and Church (CC)~\cite{Cheng2000biclustering}, where the original values are replaced with random ones to prevent them from being part of any further bicluster.

%%%%%%%%%%%%%%%%%%%%%%%%%%%%%%%%%%%%%%%%%%%%%%%%%
\subsection{Genetic algorithms}

\emph{Genetic algorithm} (GA), developed by Holland in the early 1970s~\cite{Holland1975}, emulates the principle of evolution by natural selection stated by Charles Darwin. Several biclustering approaches have been proposed based on GAs, which sometimes are referred to as evolutionary algorithms (EAs). Next, we summarize these approaches.

Bleuler et al.~\cite{Bleuler2004} proposed the first evolutionary biclustering algorithm in 2004, namely, \textbf{HEA}. This algorithm uses a binary encoding of fixed length, and the MSR criterion is used as the fitness function. In addition, a bit mutation and uniform crossover are used as variation operators to generate new biclustering solutions during the evolutionary process. In HEA, a diversity maintenance strategy is considered, which decreases the overlapping level among bicluster, and the CC algorithm~\cite{Cheng2000biclustering} is also applied as a local search method to increase the size of the biclusters. In the end, the entire population of individuals is returned as the set of resulting biclusters. Gallo et al.~\cite{Gallo2009} proposed the \textbf{BiHEA} algorithm, similar to Bleuler's approach as both perform a local search based on the CC algorithm and use the same objective function. However, the approaches differ in the crossover operators (BiHEA uses a two-point crossover). Additionally, BiHEA considers an external archive to keep the best-generated biclusters through the evolutionary process.

Divina and Aguilar-Ruiz~\cite{Divina2006} presented a sequential evolutionary biclustering algorithm (\textbf{SEBI}). The term sequential refers to that the evolutionary algorithm generates only one bicluster per run; thus, in order to generate several biclusters, SEBI needs to be invoked iteratively. Furthermore, a general matrix of weights is considered to control the overlapping among biclusters. In SEBI, three crossover and mutation strategies are used with equal probability of reproduction: one-point, two-points and uniform crossovers; and mutations that add a row or a column to the bicluster, or the standard mutation. The evaluation of individuals is carried out by an objective function that involves three different criteria: MSR, bicluster size, and row variance.

Huang et al.~\cite{QinghuaHuang2012} proposed a biclustering approach based on genetic algorithms and hierarchical clustering, called \textbf{CBEB}. First, the rows of the data matrix (conditions) are separated into a number of condition subsets (subspaces). Next, the genetic algorithm is applied to each subspace in parallel. Then, an expanding-merging strategy is employed to combine the subspaces results into output biclusters. In CBEB, the MSR metric is used as an objective function, whereas a simple crossover and a binary mutation are used to reproduce new solutions. Although this approach outperforms several traditional biclustering algorithms, it requires a longer computation time than the other methods. This disadvantage of CBEB is mainly due to its utilization process and separation method for creating and evaluating several subspaces.

An evolutionary biclustering algorithm (\textbf{EvoBic}) with a variable-length representation was proposed by Ayadi et al.~in~\cite{Ayadi2012}. This integer encoding represents the individuals as a string composed of ordered genes and conditions indices, reducing time and the memory space~\cite{DeCastro2007}. EvoBic algorithm considers three different biclustering metrics (BSize, MSR, ACF), a single-point crossover, and a standard mutation to generate new biclusters.

Another evolutionary biclustering algorithm, called \textbf{Evo-Bexpa}, was presented by Pontes et al.~\cite{Pontes2013}. This algorithm allows identifying types of biclusters in terms of different objectives. These objectives have been put together by considering a single aggregative objective function. Evo-Bexpa bases the bicluster evaluation on the use of expression patterns, recognizing both shifting and scaling patterns by considering the VE, Vol, Overlap, and Var quality metrics.

Maatouk et al.~\cite{Maatouk2014} proposed an evolutionary biclustering algorithm named \textbf{EBACross}. First, the initialization of the initial population is based on the CC algorithm~\cite{Cheng2000biclustering}. Then, the evaluation and selection of the individuals are based on four complementary biclusters metrics: bicluster size (BSize), MSR metric, average correlation (ACF), and the coefficient of variation function (CVF). In EBACross, a binary encoding of fixed length, a crossover method based on the standard deviation of the biclusters, and a mutation strategy based on the biclusters' coherence are considered. Later, the same authors proposed a generic evolutionary biclustering algorithm (\textbf{EBA})~\cite{Maatouk2019}. In this work, the authors analyzed the EBA's performance by varying its genetic components. In the study, they considered three different biclustering metrics (BSize, MSR, and ACF), two selection operators (parallel and aggregation methods), two crossover methods (random-order and biclustering methods), and two mutation operators (random and biclustering strategies). Hence, several versions of the EBA algorithm were introduced. In terms of statistical and biological significance, the clustering performance showed that the EBA configuration based on a selection with aggregation and biclustering crossover and mutation operators performed better for several microarray data. Recently, also Maatouk et al.~\cite{Maatouk2021} proposed the \textbf{ELSA} algorithm, an evolutionary algorithm based on a local search method that integrates biological information in the search process. The authors stated that statistical criteria are reflected by the size of the biclusters and the correlation between their genes, while the biological criterion is based on their biological relevance and functional enrichment degree. Thus, the ELSA algorithm evaluates the statistical and biological quality of the biclusters separately by using two objective functions based on the average correlation metric (ECF). Furthermore, in order to preserve the best biclusters over the different generations, an archiving strategy is used in ELSA.

Huang et al.~\cite{Huang2019} proposed a bi-phase evolutionary biclustering algorithm (\textbf{BP-EBA}). The first phase is dedicated to the evolution of rows and columns, and the other is for the identification of biclusters. The interaction of the two phases guides the algorithm towards feasible search directions and accelerates its convergence. BP-EBA uses a binary encoding, while the population is initialized using a hierarchical clustering strategy to discover bicluster seeds. The following biclustering metrics were employed to evaluate the individuals in the population: MSR, SMSR, BSize. Finally, the performance of this approach was compared with other biclustering algorithms using microarray datasets.

Gutierrez-Aviles \cite{Gutierrez-Aviles2014} proposed an evolutionary algorithm, \textbf{TriGen}, to find biclusters in temporal gene expression data (known as \emph{triclusters}). Thus, the aim is to find triclusters of gene expression that simultaneously take into account the experimental conditions and time points. Here, an individual is composed of three structures: a sequence of genes, a sequence of conditions, and a sequence of time points. Furthermore, the authors proposed specific genetic operators to generate new triclusters. Two different metrics were taken into account to evaluate the individual: MSR$_{\textrm{time}}$ (modification of the MSR metric) and LSL$_{\textrm{time}}$ (least-squares approximation for the points in a 3D-space representing a tricluster). As a result, TriGen could extract groups of genes with similar patterns in subsets of conditions and times, and these groups have shown to be related in terms of their functional annotations extracted from the Gene Ontology.

%%%%%%%%%%%%%%%%%%%%%%%%%%%%%%%%%%%%%%%%%%%%%%%%%
\subsection{Scatter search}

\emph{Scatter search} (SS) is a population-based evolutionary metaheuristic that emphasizes systematic processes against random procedures~\cite{Laguna2006}. The optimization process consists of evolving a set called \emph{reference}, which iteratively is updated by using \emph{combination} and \emph{improvement} methods that exploit context knowledge. In contrast to other evolutionary approaches, SS is founded on the premise that systematic designs and methods for creating new solutions afford significant benefits beyond those derived randomly.

Nepomuceno and his collaborators have developed a series of SS-based biclustering algorithms for gene expression data~\cite{Nepomuceno2011,Nepomuceno2015,Nepomuceno2015a,Nepomuceno2018}. In \cite{Nepomuceno2011}, the authors presented the \textbf{SSB} algorithm to find shifting and scaling patterns biclusters, which are interesting and relevant patterns from a biological point of view. For this purpose, the average correlation function (ACF) was modified and used as the objective function. The SSB algorithm uses a binary encoding to represent biclustering solutions and includes a local search method to improve biclusters with positively correlated genes. The same authors proposed another SS-based biclustering algorithm that integrates prior biological knowledge~\cite{Nepomuceno2015}. This algorithm (herein referred to as \textbf{SSB-bio}) requires as input, in addition to the gene expression data matrix, an annotation file that relates each gene to a set of terms from the repository \emph{Gene Ontology} (GO). Thus, two biological measures, \emph{FracGO} and \emph{SimNTO}, were proposed and integrated as part of the objective function to be optimized. This fitness function is a weighted-sum function composed of three factors: the bicluster size (BSize), the bicluster correlation (ACF metric), and the bicluster biological relevance based on FracGO or SimNTO measures (ACF$_{\textrm{bio}}$ metric).

Nepomuceno et al.~\cite{Nepomuceno2015a} proposed the \textbf{BISS} algorithm to find biclusters with both shifting and scaling patterns and negatively correlated patterns. This algorithm, similar to SSB-Bio, is based on \textit{a priori} biological information from the GO repository, particularly the categories: biological process, cellular components, and molecular function. In BISS, a fitness function involving the AFC and BSize metrics was used to evaluate the quality of the biclusters. Furthermore, in another recent study, Nepomuceno et al.~\cite{Nepomuceno2018} used the BISS algorithm for biclustering of high-dimensional expression data. This algorithm, referred here as \textbf{BISS-go}, also considers the biological knowledge available in the GO repository to find biclusters composed of groups of genes functionally coherent. This task is achieved by defining two GO semantic similarity measures integrated into the fitness function optimized by the BISS-go algorithm. The reported results showed that the inclusion of biological information improves the performance of the biclustering process.

%%%%%%%%%%%%%%%%%%%%%%%%%%%%%%%%%%%%%%%%%%%%%%%%%
\subsection{Cuckoo search}

\emph{Cuckoo search} (CS), developed by Xin-She Yang and Suash Deb~\cite{Yang2014}, is a nature-inspired metaheuristic based on the brood parasitism of some cuckoo species. In this algorithm, the exploration of the search space is enhanced using L\'{e}vy flights (L\'{e}vy distribution) instead of using simple isotropic random walks.

Yin Lu et al.~\cite{Lu2015} introduced a CS-based biclustering algorithm for gene expression data named \textbf{COCSB}. The authors incorporated different strategies in COCSB to improve its diversity performance and convergence rate, such as the searching- and abandoning-nest operations. In CBEB, the MSR metric is used as the objective function, whereas a binary bicluster representation of the solutions is considered. This approach was compared to several classical biclustering algorithms, including CC and SEBI, obtaining a good biological significance and time computation performance.

%%%%%%%%%%%%%%%%%%%%%%%%%%%%%%%%%%%%%%%%%%%%%%%%%
\subsection{Particle swarm optimization}

\emph{Particle swarm optimization} (PSO), introduced by Kennedy and Eberhart~\cite{Kennedy1995}, is a population-based search method in which the individuals (referred to as particles) are grouped into a swarm. The particles explore the search space by adjusting their trajectories iteratively according to self-experience and neighboring  particles~\cite{Laguna2006}.

Rathipriya et al.~\cite{Rathipriya2011} proposed a PSO-based biclustering algorithm called \textbf{BPSO}. BPSO was applied on web data to find biclustering that contains relationships between web users and webpages, useful for E-Commerce applications like web advertising and marketing. The individuals were encoded using the traditional binary bicluster representation, and the average correlation value (ACV) metric was used as the fitness function. This PSO-based algorithm outperformed two traditional biclustering algorithms based on greedy search. Furthermore, the identified biclusters by BPSO covered a more considerable percentage of users and webpages, capturing the global browsing patterns from web usage data.

%%%%%%%%%%%%%%%%%%%%%%%%%%%%%%%%%%%%%%%%%%%%%%%%%
\subsection{Hybrid metaheuristic approaches}

It is well-known that nature-inspired metaheuristics are effective strategies for solving optimization problems. However, sometimes it is difficult to choose a metaheuristic for a particular instance problem. In these scenarios, hybrid approaches provide flexible tools that can help to cope with this problem. In line with this, different hybrid nature-inspired metaheuristics for the biclustering problem are described below.

Nepomuceno et al.~\cite{Nepomuceno2009,Nepomuceno2010} proposed a couple of hybrid metaheuristics for biclustering based on \emph{scatter search} (SS) and \emph{genetic algorithms} (GAs). In~\cite{Nepomuceno2009}, the authors proposed the \textbf{SS\&GA} biclustering algorithm, in which the general scheme is based on SS but incorporates some GA's features such as the mutation and crossover operators to generate new biclustering solutions. This algorithm uses a binary encoding to represent solutions and considers the MSR metric to evaluate the quality of the biclusters. Later, the authors proposed a similar hybrid evolutionary algorithm (herein referred to as \textbf{HEAB}) for biclustering gene expression data~\cite{Nepomuceno2010}. In HEAB, the ACF metric is used as the fitness function based on a correlation measure. First, HEAB searches for biclusters with groups of highly correlated genes; then, new biclusters with shifting and scaling patterns are created by analyzing the correlation matrix. The experimental results using the \emph{Lymphoma dataset} indicated that the correlation-based metric outperformed the well-known MSR metric.

Recently, Lu Yin et~al.~\cite{Yin2018} proposed a hybrid biclustering approach based on \emph{cuckoo search}~(CS) and \emph{genetic algorithms}~(GAs), \textbf{GACSB}. This approach considers the CS algorithm as the main framework and uses the tournament strategy and the elite-retention strategy based on the GA to generate the next generation of solutions. In addition, GACSB uses as objective functions different metrics, namely ACV, MSR, and VE. The experimental results obtained by GACSB were compared with several classic biclustering algorithms, such as the CC algorithm and SEBI, where GACSB outperformed these algorithms when considering various gene expression datasets.

Furthermore, hybrid biclustering approaches that combine \emph{particle swarm optimization}~(PSO) features and \emph{simulated annealing}~(SA) have been proposed in the literature~\cite{Thangavel2012,Narmadha2021}. First, Thangavel et~al.~\cite{Thangavel2012} proposed a PSO-SA biclustering algorithm (\textbf{PSO-SA-BIC}) to extract biclusters of gene expression data. In this approach, SA is used as a local search procedure to improve the position of the particles with low performance. A modified version of the ACV metric is used to identify biclusters with shifting and scaling patterns. The experimental results showed that the PSO-SA-BIC algorithm outperformed some classical algorithms by providing statistically significant biclusters. Recently, Narmadha and Rathipriya~\cite{Narmadha2021} developed a hybrid approach combining PSO and SA to extract triclusters from a 3D-gene expression dataset (\emph{Yeast Cell Cycle} data). This algorithm named \textbf{HPSO-TriC} uses a fitness function based on the ACF metric, which aims to identify tricluster with a high correlation degree among genes over samples and time points (this function is referred to as ACF$_{\textrm{time}}$). The HPSO-TriC algorithm was compared with a PSO-based biclustering algorithm, performing better as the extracted tricluster was more biologically significant.

%%%%%%%%%%%%%%%%%%%%%%%%%%%%%%%%%%%%%%%%%%%%%%%%%
\subsection{Summary}

We described 23 single-objective biclustering methods based on nature-inspired metaheuristics, including simulated annealing (SA), genetic algorithm (GA), particle swarm optimization (PSO), scatter search (SS), and cuckoo search (CS). These algorithms are summarized in Table~\ref{tab:single-objective-algs}, where GA-based biclustering algorithms represent 61\% of the surveyed methods. Moreover, some hybrid approaches (combination of different metaheuristics) have been proposed due to the complexity of the biclustering problem; indeed, several of the reviewed approaches often incorporate a local search strategy to better exploit the search space. Regarding the objective function, many biclustering algorithms (43\%) combine information from multiple bicluster metrics in such a way that these optimization functions usually consider the homogeneity (e.g., the MSR metric) and the bicluster size (BSize). For measuring the bicluster homogeneity, the mean squared residence (MSR) and the average correlation function (ACF) are commonly used with percentages of 52\% and 43\%, respectively. Finally, we noticed that the binary bicluster encoding (BBE) is the most used among the biclustering algorithms (87\%), even though it is less efficient than the integer encoding (IBE) in terms of computation time and memory space.

%%%%%%%%%%%%%%%%%%%%%%%%%%%%%%%%%%%%%%%%%%%%%%%%%
%%%%%%%%%%%%%%%%%%%%%%%%%%%%%%%%%%%%%%%%%%%%%%%%%
%%%%%%%%%%%%%%%%%%%%%%%%%%%%%%%%%%%%%%%%%%%%%%%%%
\section{Multi-objective biclustering~\label{sec:multi-objective-biclustering}}

Several real-world optimization problems naturally involve multiple objectives. As the name suggests, a multi-objective optimization problem (MOOP) has a number of objective functions to be minimized or maximized. Therefore, the optimal solution for MOOPs is not a single solution but a set of solutions denoted as Pareto-optimal solutions. In this sense, a solution is Pareto-optimal if it is impossible to improve a given objective without deteriorating another. Generally, such a set of solutions represents the compromise solutions between different conflicting objectives~\cite{Deb2002}.

The biclustering problem can be formulated as a combinatorial optimization problem, such that multiple bicluster quality criteria can be optimized simultaneously~\cite{madeira2004biclustering,Seridi2015}. Indeed, in gene expression data analysis, the quality of a bicluster can be defined by its size and its intra-cluster variance (coherence). However, these criteria are independent and notably in conflict as the bicluster's coherence can constantly be improved by removing a row or a column, i.e., by reducing the bicluster's size.

This section presents different multi-objective nature-inspired metaheuristics that have been proposed for the biclustering problem. Table~\ref{tab:multi-objective-algs} summarizes some relevant details about this type of algorithms that are introduced in this section.

%%%%%%%%%%%%%%%%%%%%%%%%
\begin{table}[t!]
\small %\scriptsize %\small
\def\arraystretch{1.2}
\setlength{\tabcolsep}{0.01cm}

\caption{\label{tab:multi-objective-algs}List of multi-objective biclustering algorithms based on nature-inspired metaheuristics. \textquotedblleft MOEA\textquotedblright{} indicates the type of multi-objective evolutionary algorithm that is used as the underlying optimization strategy. \textquotedblleft Objective Functions\textquotedblright{} refers to the bicluster evaluation metrics as defined in Section~\ref{subsec:objective-functions}; and \textquotedblleft Encoding\textquotedblright{} indicates the type of bicluster encoding binary (BBE) or integer (IBE) as described in Section~\ref{subsec:Bicluster-encoding}.}

\begin{tabular*}{1\textwidth}{@{\extracolsep{\fill}}rllclr}
\toprule 
\multirow{1}{*}{\textbf{Year}} & \textbf{Algorithm} & \textbf{MOEA} & \textbf{Objective Functions} & \textbf{Encoding} & \textbf{Ref.}\tabularnewline
\midrule
2006 & \texttt{MOEAB} & NSGA-II & BSize, MSR & BBE & \cite{Mitra2006}\tabularnewline
2007 & \texttt{SMOB} & NSGA-II & BSize, MSR, rVAR & BBE & \cite{Divina2007}\tabularnewline
2008 & \texttt{MOFB} & NSGA-II & BSize, MSR, rVAR & IBE & \cite{Maulik2008}\tabularnewline
2008 & \texttt{MOPSOB} & MOPSO & BSize, MSR, rVAR & BBE & \cite{Liu2008}\tabularnewline
2008 & \texttt{CMOPSOB} & MOPSO & BSize, MSR, rVAR & BBE & \cite{Liu2009}\tabularnewline
2009 & \texttt{HMOPSOB} & MOPSO & BSize, MSR, rVAR & BBE & \cite{Lashkargir2009}\tabularnewline
2009 & \texttt{MOACOB} & MOACO & BSize, MSR & BBE & \cite{Liu2009b}\tabularnewline
2009 & \texttt{MOM-aiNet} & MOAIS & BSize, MSR & IBE & \cite{Coelho2009}\tabularnewline
2009 & \texttt{MOGAB} & NSGA-II & MSR, rVAR & IBE & \cite{Maulik2009}\tabularnewline
2009 & \texttt{SPEA2B} & SPEA2 & BSize, MSR, rVAR  & BBE & \cite{Gallo2009a}\tabularnewline
2011 & \texttt{MOBI} & NSGA-II & BSize, MSR, rVAR & IBE & \cite{Seridi2011}\tabularnewline
2012 & \texttt{SMOB-VE} & NSGA-II & BSize, VE, rVAR & BBE & \cite{Divina2012}\tabularnewline
2015 & \texttt{HMOBI} & IBEA & MSR, rVAR & IBE & \cite{Seridi2015}\tabularnewline
2015 & \texttt{SPEA2B}-$\delta$ & SPEA2 & BSize, MSR  & BBE & \cite{Golchin2015}\tabularnewline
2016 & \texttt{AMOSAB} & AMOSA & BSize, MSR & IBE & \cite{Sahoo2016}\tabularnewline
2017 & \texttt{PBD-SPEA2} & SPEA2 & BSize, MSR, rVAR  & IBE$^{*}$ & \cite{Golchin2017}\tabularnewline
2019 & \texttt{BP-NSGA2} & NSGA-II & BSize, MSR & BBE & \cite{Kong2019}\tabularnewline
2019 & \texttt{AMOSAB} & AMOSA & BSize, MSR & IBE & \cite{Acharya2019}\tabularnewline
2020 & \texttt{MMCo-Clus} & NSGA-II & MSR, rVAR, AI & BBE$^{*}$ & \cite{Cui2020}\tabularnewline
\bottomrule
\end{tabular*}
$\left(*\right)$ Modification to the original encoding described in Section~\ref{subsec:Bicluster-encoding}.

\end{table}
%%%%%%%%%%%%%%%%%%%%%%%%

%%%%%%%%%%%%%%%%%%%%%%%%%%%%%%%%%%%%%%%%%%%%%%%%%
\subsection{Multi-objective evolutionary algorithms based on \texttt{NSGA-II}}

One of the most representative multi-objective algorithms is the non-dominated sorting genetic algorithm (NSGA-II)~\cite{Deb2002}. This Pareto-dominance algorithm is characterized by incorporating an explicit diversity-preservation mechanism. As NSGA-II has widely been used to address the biclustering problem, we will outline how the algorithm operates. In NSGA-II, the offspring population $Q_{t}$ is first created by using the parent population $P_{t}$; then, the two populations are combined to form the population $R_{t}$. Next, a non-dominated sorting method is used to classify the entire population $R_{t}$. Then, the new population $P_{t+1}$ is filled by solutions of different non-dominated fronts, starting with the best non-dominated front, followed by the second front, and so on. Finally, when the last allowed front is being considered,  there may exist more solutions than the remaining slots in the new population; in this case, a niching strategy based on the crowding distance is used to choose the members of the last front. For a more detailed description and understanding of NSGA-II, the reader is referred to~\cite{Deb2002}. Next, we describe different biclustering algorithms that use NSGA-II as the underlying optimization method.

Several multi-objective biclustering approaches have been proposed based on the well-known NSGA-II algorithm~\cite{Mitra2006,Divina2007,Maulik2008,Maulik2009,Seridi2011,Divina2012,Kong2019,Cui2020}. Research in multi-objective biclustering became popular after the work by Mitra and Banka~\cite{Mitra2006} entitled \textquotedbl Multi-objective evolutionary biclustering of gene expression data,\textquotedbl{} which was published in 2006. This algorithm (referred to as \textbf{MOEAB}) uses the CC method~\cite{Cheng2000biclustering} during the population's initialization and after applying the variation operators. MOEAb uses a binary encoding representation (BBE), a uniform single-point crossover, and a single-bit mutation. Concerning the objective functions, the size of the bicluster (BSize) and the MSR metric were considered.

Divina et al.~\cite{Divina2007,Divina2012} have proposed some multi-objective biclustering approaches based on NSGA-II for microarray data. First, Divina and Aguilar-Ruiz~\cite{Divina2007} presented the sequential multi-objective biclustering (\textbf{SMOB}) algorithm for finding biclusters of high quality with large variation. SMOB adopts a sequential strategy such that the algorithm is invoked several times, each time returning a bicluster stored in a temporal list. This algorithm considered a binary encoding, three different crossover operators (one-point, two-point, and uniform), and tree mutation strategies (single-bit, add-row, and add-column). The objectives considered in SMOB were: the MSR metric, the bicluster size (BSize), and the row variance (rVAR). Later, Divina et al.~\cite{Divina2012} presented the virtual error (VE) metric, which measures how well the genes in a bicluster follow the general tendency. Then, the VE metric was used as an additional objective function in SNOB. This modified algorithm, referred to as \textbf{SMOB-VE}, aims to find biclusters with shifting and scaling patterns using VE instead of the MSR metric.

Maulik et al.~\cite{Maulik2008,Maulik2009} also presented biclustering algorithms based on the NSGA-II algorithm. In~\cite{Maulik2008}, the authors presented a multi-objective fuzzy biclustering (\textbf{MOFB}) algorithm for discovering overlapping biclusters. MOFB simultaneously optimizes fuzzy versions of the metrics MSR, BSize, and rVAR. Furthermore, MOFB uses an integer encoding of variable string length, a single-point crossover, and a uniform mutation strategy. Subsequently, the authors proposed the \textbf{MOGAB} algorithm~\cite{Maulik2009}, which optimizes two objective functions, MSR and rVAR. Similar to MOFB, the MOGAB algorithm uses a variable string length encoding, a single-point crossover, and a uniform mutation strategy. Additionally, the authors presented the bicluster index (BI) to validate the obtained biclusters from microarray data.

Seridi et al.~\cite{Seridi2011} proposed a multi-objective biclustering algorithm that simultaneously optimizes three conflicting objectives, the bicluster size (BSize), the MSR metric, and the row variance (rVAR). The presented evolutionary framework \textbf{MOBI} can integrate any evolutionary algorithm, such as NSGA-II in this case. MOBI uses an integer bicluster encoding, a single-point crossover, and the CC local-search heuristic~\cite{Cheng2000biclustering} replaces the mutation operator.

Kong et al.~\cite{Kong2019} presented an interesting biclustering algorithm incorporating a bi-phase evolutionary architecture and the NSGA-II algorithm. In this algorithm, referred to as \textbf{BP-NSGA2}, the first phase consists of evolving the population of rows and columns and then the population of biclusters. The two populations are initialized using a hierarchical clustering method, and then they are evolved independently. Next, during the evolutionary process, the NSGA-II algorithm optimizes the MSR metric and bicluster size simultaneously. This multi-objective approach outperformed two traditional biclustering approaches. The same authors revisited this idea of incorporating a bi-phase evolutionary strategy for the proposal of the BP-EBA algorithm~\cite{Huang2019}.

Recently, Cui et al.~\cite{Cui2020} proposed a multi-objective optimization-based multi-view co-clustering algorithm (named \textbf{MMCo-Clus}) for feature selection of gene expression data. First, two data views are constructed using information from two different biological data sources. Next, the MMCo-Clus algorithm identifies biclusters (co-clustering solutions) considering the constructed views. Finally, a small number of non-redundant features are selected from the original feature space using consensus clustering. MMCo-Clus uses two well-known bicluster measures, the MSR metric and the bicluster size (BSize), and the agreement index. Although this approach focuses on feature selection, it applies an intrinsic biclustering strategy to select relevant features.

%%%%%%%%%%%%%%%%%%%%%%%%%%%%%%%%%%%%%%%%%%%%%%%%%
\subsection{Multi-objective evolutionary algorithms based on \texttt{SPEA2} and \texttt{IBEA}}

Similar to the NSGA-II algorithm, the strength Pareto evolutionary algorithm (SPEA2)~\cite{Zitzler2001} is a well-known multi-objective evolutionary algorithm (MOEA) in the specialized literature. This algorithm is characterized by maintaining an external population, which is used to introduce elitism. This population stores a fixed number of non-dominated solutions founded during the entire evolutionary process. Additionally, SPEA2 uses these elite solutions to participate in the genetic operations along with the current population to improve the convergence and diversity of the algorithm. Below we describe some biclustering approaches that use SPEA2 as the primary multi-objective optimization method.

Gallo et al.~\cite{Gallo2009a} addressed the microarray biclustering problem using different MOEAs, where the SPEA2 algorithm performed better. This approach (\textbf{SPEA2B}) considered a binary encoding, a probabilistic-based mutation, and a two-point crossover operator. Four different objectives were considered in SPEA2B: the number of genes, number of conditions, row variance (rVAR), and the MSR metric. Additionally, a greedy method was implemented based on the CC algorithm to maintain large size and low homogeneity biclusters in the population.

Golchin et al.~\cite{Golchin2015,Golchin2017} have proposed biclustering approaches based on the SPEA2 algorithm. First, Golchin et al.~\cite{Golchin2015} presented a multi-objective biclustering algorithm (herein referred to as \textbf{SPEA2B-$\delta$}), which optimizes the MSR metric and the size of the bicluster simultaneously. This algorithm used a binary bicluster encoding, a single-point crossover, and a single-bit mutation operator. SPEA2B-$\delta$ also incorporated a search heuristic strategy similar to the CC algorithm to remove unwanted genes and conditions. As SPEA2B-$\delta$ generates a set of biclustering solutions (Pareto front approximation), a fitness selection function based on the coherence and size of the biclusters is considered to choose the best solutions. Later on, Golchin and Liew~\cite{Golchin2017} proposed a SPEA2-based biclustering algorithm for gene expression data named \textbf{PBD-SPEA2}. This algorithm considers three objective functions, namely MSR, BSize, and rVAR. An interesting aspect in PBD-SPEA2 is that each individual in the population represents multiple bicluster, instead of only one bicluster as in other approaches. Thus, given a user-defined number of biclusters, $k$, an integer-based encoding of fixed length but extended to $k$ biclusters is considered. Regarding the variation operators for generating new solutions, PBD-SPEA2 uses the CC heuristic as a mutation operator and a similarity-based crossover. Finally, a sequential selection technique is used to choose the final solution from the Pareto from approximations obtained by PBD-SPEA2.

Similar to NSGA-II and SPEA2 algorithms, the indicator-based multi-objective algorithm (IBEA)~\cite{Zitzler2004} is another representative MOEA in the evolutionary computation literature. In this regard, Seridi et al.~\cite{Seridi2015} proposed a biclustering approach based on the IBEA algorithm named \textbf{HMOBI}~\cite{Seridi2015}. This algorithm used an integer-based representation, a single-point crossover, and a mutation operator based on the CC algorithm. In HMOBI, three biclustering metrics are considered as objective functions, the MSR metric and the row variance, and the bicluster size. The obtained results were compared in terms of the bicluster quality and their biological relevance.

%%%%%%%%%%%%%%%%%%%%%%%%%%%%%%%%%%%%%%%%%%%%%%%%%
\subsection{Multi-objective particle swarm optimization}

The particle swarm optimization (PSO) algorithm has been extended to solve multi-objective optimization problems (MOPs). Most of the existing multi-objective PSO (MOPSO) algorithms involve developments from the evolutionary computation field to address MOPs. For a review of different MOPSO algorithms, the reader is referred to the survey by Reyes-Sierra~\cite{Reyes2006}. Below we present different MOPSO algorithms for the biclustering problem.

Two multi-objective approaches to microarray biclustering based on MOPSO algorithms were presented by Junwan Liu et al.~\cite{Liu2008,Liu2009}. Both approaches use a binary bicluster encoding and optimize simultaneously three objective functions: the bicluster size (BSize), the MSR metric, and the row variance (rVAR). The first algorithm, named \textbf{MOPSOB}~\cite{Liu2008}, considers a relaxed form of the Pareto dominance ($\in$-dominance), whereas the second approach (\textbf{CMOPSOB}~\cite{Liu2009}) uses a Pareto-based dominance as in the NSGA-II algorithm. Additionally, in the CMOPSOB algorithm, the information of nearest neighbors between particles is considered when updating the particles' velocity, aiming to accelerate the algorithm's convergence. In the comparative analysis, the biological relevance of the biclusters obtained by CMOPSOB was analyzed considering the information of the GO repository, showing that this approach was able to find biologically meaningful clusters.

Another gene expression biclustering algorithm based on a MOPSO was proposed by Lashkargir et al.~\cite{Lashkargir2009}. The authors proposed a hybrid MOPSO for biclustering algorithm (\textbf{HMOPSOB}), which uses a binary encoding and optimizes four objective functions: bicluster size, row variance, and the MSR metric. The HMOPSOB algorithm includes a local search method based on the CC algorithm and, in addition to the PSO steps, three mutation operators: standard, add-row, and add-column. Additionally, this approach can find biclusters with a low level of overlap among biclusters by considering an external archive.

%%%%%%%%%%%%%%%%%%%%%%%%%%%%%%%%%%%%%%%%%%%%%%%%%
\subsection{Other multi-objective approaches }

This section describes a number of multi-objective approaches to biclustering that do not fit into the previous classification. The approaches described below are based on ant colony optimization (ACO), artificial immune systems (AISs), and simulated annealing (SA).

The ACO algorithm~\cite{Dorigo2005} is a probabilistic technique inspired by the behavior of ants in finding paths from their colony to a food source. In this regard, a multi-objective ACO algorithm for microarray biclustering (\textbf{MOACOB}) was introduced by Junwan Liu et al.~\cite{Liu2009b}. MOACOB algorithm uses ACO concepts for biclustering microarray data, where the bicluster size and the MSR metric are optimized simultaneously. Furthermore, this algorithm uses  a relaxed form of the Pareto dominance ($\in$-dominance) and considers a binary encoding to represent biclusters. In MOACOB, a number of ants probabilistically construct solutions using a given pheromone model; then, a local search procedure is applied to the constructed solutions. In general, this multi-objective approach based on ACO outperformed another three biclustering algorithms in terms of the size of the obtained biclusters.

AISs are inspired by the principles of immunology and the observed immune process of vertebrates~\cite{Talbi2009}. Additionally, AIS is highly robust, adaptive, inherently parallel, and self-organized. In line with these, Coelho et al.~\cite{Coelho2009} proposed a multi-objective biclustering algorithm based on AIS to analyze texts, named \textbf{BIC-aiNet}. In the text mining problem, the input data matrix is composed of rows (texts) and columns (attributes of the corresponding texts), and the aim is to find bipartitions of the whole dataset. The BIC-aiNet algorithm uses an integer bicluster representation, a mutation strategy to insert or remove rows and columns, and a suppression procedure to eliminate entire biclusters if a particular condition is satisfied. This approach was compared with the k-means algorithm, showing that BIC-aiNet discovered more meaningful text biclusters.

The simulated annealing-based multi-objective optimization algorithm (AMOSA)~\cite{Bandyopadhyay2008a} has been used as the underlying optimization strategy for finding bicluster in gene expression data~\cite{Acharya2019,Sahoo2016}. First, Sahoo et al.~\cite{Sahoo2016} presented an AMOSA-based biclustering algorithm (\textbf{AMOSAB}) that optimized the MSR metric and the row variance (rVAR) simultaneously. AMOSAB used a real-based encoding of biclusters and a decodification method based on the Euclidean distance to obtain the final biclusters. Then, the same authors, Acharya et al.~\cite{Acharya2019}, proposed modifications to the AMOSAB algorithm where the decodification method considered three different distance functions: Euclidean, Point Symmetry (PS), and Line Symmetry (LS). The results showed that the AMOSAB algorithm using the PS and LS distance performed better than the Euclidean version.

%%%%%%%%%%%%%%%%%%%%%%%%%%%%%%%%%%%%%%%%%%%%%%%%%
\subsection{Summary}

We analyzed 19 biclustering approaches that use different nature-inspired multi-objective metaheuristics such as NSGA-II, SPEA2, MOPSO, and AMOSA. The main characteristics of these methods are summarized in Table~\ref{tab:multi-objective-algs}, where multi-objective evolutionary algorithms are the most widely used with 63\%, the NSGA-II algorithm being the most common of this group with 42\%. Regarding the objective functions used to guide the search of the multi-objective algorithms, we noticed that the bicluster size (BSize), the bicluster coherence (MSR metric), and the row variance (rVAR) are commonly optimized simultaneously as these criteria are independent and usually in conflict. Finally, regarding the types of bicluster representations, it was noted that both techniques are used almost equally, the binary representation (BBE) with 58\%, while the integer representation (IBE) with 42\%. However, in multi-objective algorithms, there is a tendency in recent years to use the IBE representation, which is more efficient than the binary representation for biclustering problems.

%%%%%%%%%%%%%%%%%%%%%%%%%%%%%%%%%%%%%%%%%%%%%%%%%
%%%%%%%%%%%%%%%%%%%%%%%%%%%%%%%%%%%%%%%%%%%%%%%%%
%%%%%%%%%%%%%%%%%%%%%%%%%%%%%%%%%%%%%%%%%%%%%%%%%
\section{Discussion and future directions~\label{sec:discussion}}

%%%%%%%%%%%%%%%%%%%%%%%%%
\begin{figure}[!t]
\begin{centering}
\includegraphics[width=1\textwidth]{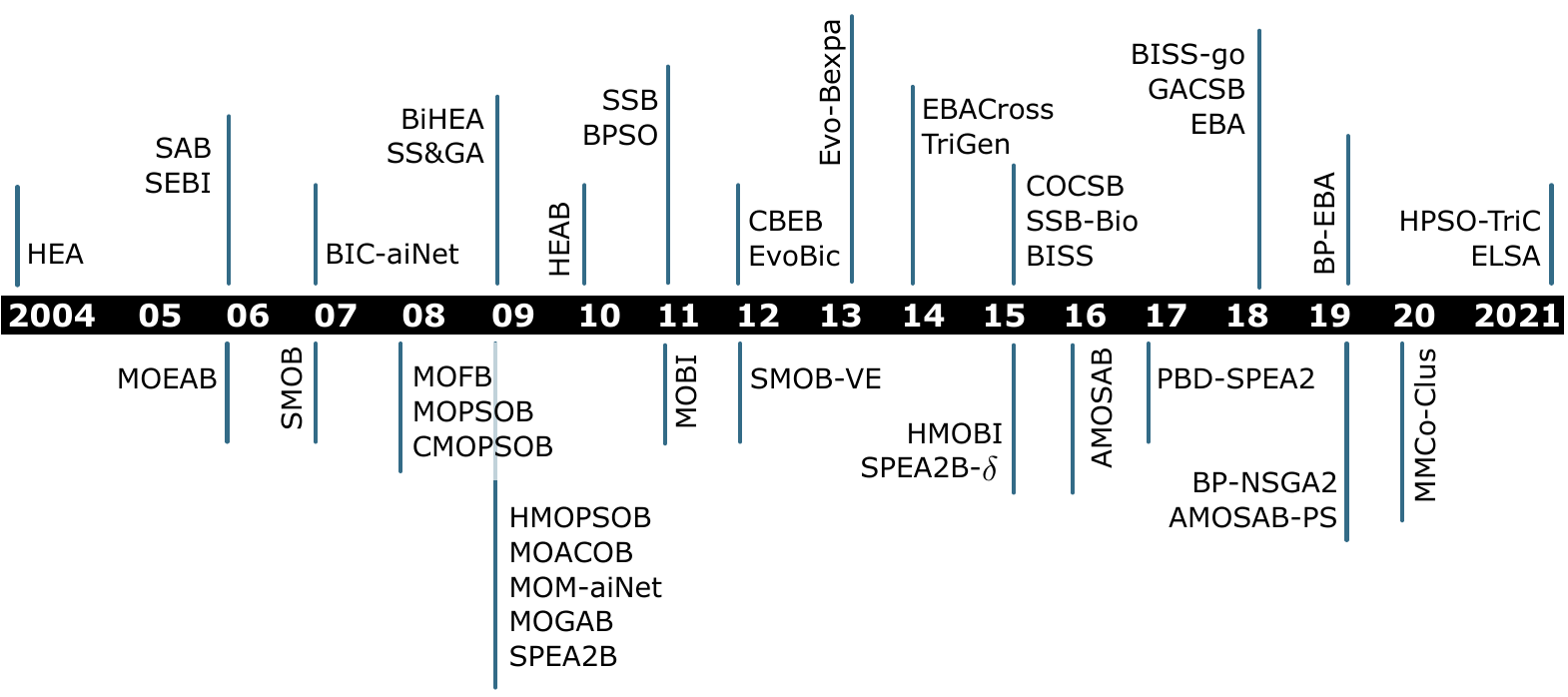}
\par\end{centering}
\caption{\label{fig:timeline-algorithms}Timeline of biclustering algorithms based on nature-inspired metaheuristics. Above the timeline, the 23 single-objective approaches are presented, whereas below the 19 multi-objective algorithms are illustrated.}
\end{figure}
%%%%%%%%%%%%%%%%%%%%%%%%%

Metaheuristic-based biclustering algorithms have gained much relevance in recent decades, mainly because biclustering remains an important problem in practice and, computationally, it is a highly combinatorial problem. In this sense, Figure~\ref{fig:timeline-algorithms} illustrates the 42 algorithms, surveyed in this book chapter, that have been proposed within the last 15 years.

In this chapter, we introduced the fundamental concepts and principles of biclustering optimization methods, which could be classified into two main categories depending on their underlying search strategy, namely single-objective and multi-objective biclustering approaches. Indeed, single-objective metaheuristics represent 55\% of the surveyed algorithms whereas multi-objective algorithms represent 45\%. Overall, biclustering approaches based on the evolutionary computation paradigm represent 71\% of the reviewed works; the remaining 29\% corresponds to other types of nature-inspired metaheuristics. Regarding the representation of biclusters, the binary encoding is the most used with 73\%, but there is a tendency to use the integer encoding, especially in multi-objective approaches.

Based on the revised works, it is notable that a large number of single-objective approaches (44\%) consider as objective function a combination of multiple biclustering criteria, suggesting that the biclustering problem is inherently a multi-objective optimization problem. Besides, multi-objective biclustering algorithms optimize multiple biclustering metrics simultaneously, having the advantage of discovering several types of biclusters (see Section~\ref{subsec:biclustering-definition}). However, these approaches generate a set of solutions requiring an additional mechanism to filter and select the best biclustering solution. Therefore, the optimization technique's selection depends on the complexity of the biclustering problem in terms of the type of biclusters and the bicluster structures to be discovered (overlapping among bicluster and matrix covering matrix).

\subsection*{Future directions}

Biclustering is an open field with several research directions, opportunities, and challenges that involve the following issues:

\begin{itemize} 
\item To solve complex biclustering problems to discover different bicluster types and bicluster structures, it is necessary to study and analyze the current objective function and bicluster representations that will help to select the appropriate optimization scheme and components according to the biclustering scenario. For instance, it is clear that an integer based-representation is preferable over a binary representation~\cite{Seridi2011,Acharya2019}. Most of these approaches using an integer representation encodes a single bicluster; however, it is possible to codify multiple biclusters in a single solution, as demonstrated recently by Golchin~\cite{Golchin2017}.

\item Novel nature-inspired metaheuristics are continuously proposed in the literature as potential approaches to solve the biclustering problem. Particularly, many-objective optimization evolutionary algorithms such as MOEA/D~\cite{Zhang2007}, NSGA-III~\cite{Deb2014} can be used to cope with multiple biclustering criteria (more than three objective functions). It is important to mention that selecting the best biclustering solution is an added challenge when using these multi-objective biclustering approaches.

\item Although many synthetic and real-life datasets have been used systematically in the literature, there is no recognized benchmark that the research community could use to evaluate and compare biclustering approaches. Such a benchmark should include different types of bicluster, diverse bicluster structures, noise, overlapping, etc. Furthermore, when comparing the performance of biclustering approaches, it is crucial to consider their statistical and biological significance (i.e., to consider the available biological information~\cite{Maatouk2021}).

\item Recently, the biclustering problem is referred to as triclustering when the time dimension is considered in addition to rows and columns information. Finding tricluster when considering temporal data brings up new research challenges as it will require the adaptation of current algorithms, bicluster metrics, evaluation measures, etc. Indeed, the triclustering problem has been addressed recently using single-objective metaheuristics~\cite{Gutierrez-Aviles2014,Narmadha2021}; however, there are opportunities to address this problem as multi-objective optimization.

\item Most of the proposed metaheuristic-based biclustering algorithms have been designed to work on biological data (mainly gene expression and microarray data). Thus, the application of biclustering to other data types, such as heterogeneous data, is very limited as it brings up additional challenges. In~\cite{vandromme2020biclustering}, the authors proposed a greedy procedure to extract biclusters from heterogeneous, temporal, and large-scale data. This procedure has been applied successfully on Electronic Health Records (EHR) thanks to the sparsity of data in this scenario, optimizing the enumeration. It will be interesting to study the potential of nature-inspired metaheuristics to discover heterogeneous-like bicluster in EHR applications.

\item There are many application domains where multiple pieces of information are available for each individual subject. For instance, multiple data matrices might be available for the same set of genes and conditions. In this regard, multi-view data clustering algorithms can integrate these information pieces to find consistent clusters across different data views~\cite{Jose-Garcia2019,Jose-Garcia2021}. This same multi-view clustering concept can be extended to biclustering, where the aim is to discover biclusters across multiple data matrices (i.e., data views).
\end{itemize}

%%%%%%%%%%%%%%%%%%%%%%%%%%%%%%%%%%%%%%%%%%%%%%%%%
%%%%%%%%%%%%%%%%%%%%%%%%%%%%%%%%%%%%%%%%%%%%%%%%%
%%%%%%%%%%%%%%%%%%%%%%%%%%%%%%%%%%%%%%%%%%%%%%%%%
\section{Conclusion~\label{sec:conclusion}}

Biclustering has emerged as an important approach and currently plays an essential role in various applications ranging from bioinformatics to text mining. Different nature-inspired metaheuristics have been applied to address the biclustering problems as, from the computational point of view, this is a NP-hard optimization problem.

In this regard, this chapter presented a detailed survey of metaheuristics approaches to address the biclustering problem. The review focused on the underlying optimization methods and their main search components: biclustering encoding, variation operators and bicluster objective functions. This review focused on single versus multi-objective approaches. Additionally, we presented a discussion and some emerging research directions.

%%%%%%%%%%%%%%%%%%%%%%%%%%%%%%%%%%%%%%%%%%%%%%%%%
%%%%%%%%%%%%%%%%%%%%%%%%%%%%%%%%%%%%%%%%%%%%%%%%%
%%%%%%%%%%%%%%%%%%%%%%%%%%%%%%%%%%%%%%%%%%%%%%%%%
\subsection*{Acknowledgement}

This work has been partially supported by the I-Site ULNE (Universit\'e Lille-Nord Europe) and the Lille European Metropolis (MEL).
%%%%%%%%%%%%%%%%%%%%%%%%%%%%%%%%%%%%%%%%%%%%%%%%%
%%%%%%%%%%%%%%%%%%%%%%%%%%%%%%%%%%%%%%%%%%%%%%%%%
%%%%%%%%%%%%%%%%%%%%%%%%%%%%%%%%%%%%%%%%%%%%%%%%%
\bibliographystyle{splncs04}
\bibliography{references_biclustering}

\begin{thebibliography}{10}
\providecommand{\url}[1]{\texttt{#1}}
\providecommand{\urlprefix}{URL }
\providecommand{\doi}[1]{https://doi.org/#1}

\bibitem{Acharya2019}
Acharya, S., Saha, S., Sahoo, P.: {Bi-clustering of microarray data using a
  symmetry-based multi-objective optimization framework}. Soft Computing
  \textbf{23}(14),  5693--5714 (2019). \doi{10.1007/s00500-018-3227-5}

\bibitem{aguilar2005shifting}
Aguilar-Ruiz, J.S.: Shifting and scaling patterns from gene expression data.
  Bioinformatics  \textbf{21}(20),  3840--3845 (2005)

\bibitem{Ayadi2012}
Ayadi, W., Maatouk, O., Bouziri, H.: {Evolutionary biclustering algorithm of
  gene expression data}. In: International Workshop on Database and Expert
  Systems Applications. pp. 206--210. IEEE (2012). \doi{10.1109/DEXA.2012.46}

\bibitem{Bandyopadhyay2008a}
Bandyopadhyay, S., Saha, S., Maulik, U., Deb, K.: {A simulated annealing-based
  multiobjective optimization algorithm: AMOSA}. IEEE Transactions on
  Evolutionary Computation  \textbf{12}(3),  269--283 (2008).
  \doi{10.1109/TEVC.2007.900837}

\bibitem{Ben-Dor2003}
Ben-Dor, A., Chor, B., Karp, R., Yakhini, Z.: {Discovering local structure in
  gene expression data: the order-preserving submatrix problem}. Journal of
  Computational Biology  \textbf{10}(3-4),  373--384 (jun 2003).
  \doi{10.1089/10665270360688075}

\bibitem{Bleuler2004}
Bleuler, S., Prelic, A., Zitzler, E.: {An EA framework for biclustering of gene
  expression data}. In: Congress on Evolutionary Computation (CEC'2004). pp.
  166--173. IEEE (2004). \doi{10.1109/CEC.2004.1330853}

\bibitem{Bryan2006}
Bryan, K., Cunningham, P., Bolshakova, N.: {Application of simulated annealing
  to the biclustering of gene expression data}. IEEE Transactions on
  Information Technology in Biomedicine  \textbf{10}(3),  519--525 (2006).
  \doi{10.1109/TITB.2006.872073}

\bibitem{DeCastro2007}
de~Castro, P.A.D., de~Fran{\c{c}}a, F.O., Ferreira, H.M., {Von Zuben}, F.J.:
  {Applying biclustering to text mining: an immune-inspired approach}. In:
  International Conference on Artificial Immune Systems. pp. 83--94 (2007).
  \doi{10.1007/978-3-540-73922-7\_8}

\bibitem{Cheng2000biclustering}
Cheng, Y., Church, G.M.: Biclustering of expression data. In: International
  Conference on Intelligent Systems for Molecular Biology. vol.~8, pp. 93--103
  (2000)

\bibitem{ChunTang2001}
{Chun Tang}, {Li Zhang}, {Aidong Zhang}, Ramanathan, M.: {Interrelated two-way
  clustering: an unsupervised approach for gene expression data analysis}. In:
  IEEE International Symposium on Bioinformatics and Bioengineering. pp.
  41--48. IEEE (2001). \doi{10.1109/BIBE.2001.974410}

\bibitem{Coelho2009}
Coelho, G.P., de~Fran{\c{c}}a, F.O., {Von Zuben}, F.J.: {Multi-objective
  biclustering: when non-dominated solutions are not enough}. Journal of
  Mathematical Modelling and Algorithms  \textbf{8}(2),  175--202 (2009).
  \doi{10.1007/s10852-009-9102-8}

\bibitem{Cui2020}
Cui, L., Acharya, S., Mishra, S., Pan, Y., Huang, J.Z.: {MMCO-Clus -- an
  evolutionary co-clustering algorithm for gene selection}. IEEE Transactions
  on Knowledge and Data Engineering pp.~1--1 (2020).
  \doi{10.1109/TKDE.2020.3035695}

\bibitem{Deb2014}
Deb, K., Jain, H.: {An evolutionary many-objective optimization algorithm using
  reference-point-based nondominated sorting approach, Part I: Solving problems
  with box constraints}. IEEE Transactions on Evolutionary Computation
  \textbf{18}(4),  577--601 (2014). \doi{10.1109/TEVC.2013.2281535}

\bibitem{Deb2002}
Deb, K., Pratap, A., Agarwal, S., Meyarivan, T.: {A fast and elitist
  multiobjective genetic algorithm: NSGA-II}. IEEE Transactions on Evolutionary
  Computation  \textbf{6}(2),  182--197 (2002). \doi{10.1109/4235.996017}

\bibitem{dhaenens2016metaheuristics}
Dhaenens, C., Jourdan, L.: Metaheuristics for big data. John Wiley \& Sons
  (2016)

\bibitem{dhaenens2019metaheuristics}
Dhaenens, C., Jourdan, L.: Metaheuristics for data mining. 4OR  \textbf{17}(2),
   115--139 (2019)

\bibitem{ding2006biclustering}
Ding, C., Zhang, Y., Li, T., Holbrook, S.R.: Biclustering protein complex
  interactions with a biclique finding algorithm. In: Sixth International
  Conference on Data Mining (ICDM'06). pp. 178--187. IEEE (2006)

\bibitem{Divina2006}
Divina, F., Aguilar-Ruiz, J.S.: {Biclustering of expression data with
  evolutionary computation}. IEEE Transactions on Knowledge and Data
  Engineering  \textbf{18}(5),  590--602 (2006). \doi{10.1109/TKDE.2006.74}

\bibitem{Divina2007}
Divina, F., Aguilar-Ruiz, J.S.: {A multi-objective approach to discover
  biclusters in microarray data}. In: Genetic and Evolutionary Computation
  Conference - GECCO '07. p.~385. ACM Press (2007).
  \doi{10.1145/1276958.1277038}

\bibitem{Divina2012}
Divina, F., Pontes, B., Gir{\'{a}}ldez, R., Aguilar-Ruiz, J.S.: {An effective
  measure for assessing the quality of biclusters}. Computers in Biology and
  Medicine  \textbf{42}(2),  245--256 (2012).
  \doi{10.1016/j.compbiomed.2011.11.015}

\bibitem{Dorigo2005}
Dorigo, M., Blum, C.: {Ant colony optimization theory: A survey}. Theoretical
  Computer Science  \textbf{344}(2-3),  243--278 (2005).
  \doi{10.1016/j.tcs.2005.05.020}

\bibitem{Gallo2009}
Gallo, C.A., Carballido, J.A., Ponzoni, I.: {BiHEA: A hybrid evolutionary
  approach for microarray biclustering}. In: Brazilian Symposium on
  Bioinformatics. pp. 36--47 (2009). \doi{10.1007/978-3-642-03223-3\_4}

\bibitem{Gallo2009a}
Gallo, C.A., Carballido, J.A., Ponzoni, I.: {Microarray biclustering: A novel
  memetic approach based on the PISA platform}. In: European Conference on
  Evolutionary Computation, Machine Learning and Data Mining in Bioinformatics.
  pp. 44--55 (2009). \doi{10.1007/978-3-642-01184-9\_5}

\bibitem{Getz2000}
Getz, G., Levine, E., Domany, E.: {Coupled two-way clustering analysis of gene
  microarray data}. Proceedings of the National Academy of Sciences
  \textbf{97}(22),  12079--12084 (2000). \doi{10.1073/pnas.210134797}

\bibitem{Golchin2015}
Golchin, M., Davarpanah, S.H., Liew, A.W.C.: {Biclustering analysis of gene
  expression data using multi-objective evolutionary algorithms}. In:
  International Conference on Machine Learning and Cybernetics. pp. 505--510.
  IEEE (2015). \doi{10.1109/ICMLC.2015.7340608}

\bibitem{Golchin2017}
Golchin, M., Liew, A.W.C.: {Parallel biclustering detection using strength
  Pareto front evolutionary algorithm}. Information Sciences  \textbf{415-416},
   283--297 (2017). \doi{10.1016/j.ins.2017.06.031}

\bibitem{Gu2008}
Gu, J., Liu, J.S.: {Bayesian biclustering of gene expression data}. BMC
  Genomics  \textbf{9}(Suppl 1), ~S4 (2008). \doi{10.1186/1471-2164-9-S1-S4}

\bibitem{Gutierrez-Aviles2014}
Guti{\'{e}}rrez-Avil{\'{e}}s, D., Rubio-Escudero, C.,
  Mart{\'{i}}nez-{\'{A}}lvarez, F., Riquelme, J.: {TriGen: A genetic algorithm
  to mine triclusters in temporal gene expression data}. Neurocomputing
  \textbf{132},  42--53 (2014). \doi{10.1016/j.neucom.2013.03.061}

\bibitem{Hartigan1972}
Hartigan, J.A.: {Direct clustering of a data matrix}. Journal of the American
  Statistical Association  \textbf{67}(337), ~123 (1972). \doi{10.2307/2284710}

\bibitem{Hochreiter2010}
Hochreiter, S., Bodenhofer, U., Heusel, M., Mayr, A., Mitterecker, A., Kasim,
  A., Khamiakova, T., {Van Sanden}, S., Lin, D., Talloen, W., Bijnens, L.,
  G{\"{o}}hlmann, H.W.H., Shkedy, Z., Clevert, D.A.: {FABIA: factor analysis
  for bicluster acquisition}. Bioinformatics  \textbf{26}(12),  1520--1527
  (2010). \doi{10.1093/bioinformatics/btq227}

\bibitem{Holland1975}
Holland, J.H.: {Adaptation in natural and artificial systems}. University of
  Michigan Press, Michigan, USA (1975)

\bibitem{Huang2019}
Huang, Q., Huang, X., Kong, Z., Li, X., Tao, D.: {Bi-phase evolutionary
  searching for biclusters in gene expression data}. IEEE Transactions on
  Evolutionary Computation  \textbf{23}(5),  803--814 (2019).
  \doi{10.1109/TEVC.2018.2884521}

\bibitem{Jose-Garcia2015}
Jos{\'{e}}-Garc{\'{i}}a, A., G{\'{o}}mez-Flores, W.: {Automatic clustering
  using nature-inspired metaheuristics: A survey}. Applied Soft Computing
  \textbf{41},  192--213 (2016). \doi{10.1016/j.asoc.2015.12.001}

\bibitem{Jose-Garcia2019}
Jos{\'{e}}-Garc{\'{i}}a, A., Handl, J., G{\'{o}}mez-Flores, W., Garza-Fabre,
  M.: {Many-view clustering: An illustration using multiple dissimilarity
  measures}. In: Genetic and Evolutionary Computation Conference - GECCO '19.
  pp. 213--214. ACM Press (2019). \doi{10.1145/3319619.3323365}

\bibitem{Jose-Garcia2021}
{Jos{\'{e}}-Garc{\'{i}}a}, A., Handl, J., {G{\'{o}}mez-Flores}, W.,
  {Garza-Fabre}, M.: {An evolutionary many-objective approach to multiview
  clustering using feature and relational data}. Applied Soft Computing
  \textbf{108} (2021). \doi{https://doi.org/10.1016/j.asoc.2021.107425"}

\bibitem{Kennedy1995}
Kennedy, J., Eberhart, R.: {Particle Swarm Optimization}. In: International
  Conference on Neural Networks. vol.~4, pp. 1942--1948. IEEE (1995).
  \doi{10.1109/ICNN.1995.488968}

\bibitem{Kirkpatrick1983}
Kirkpatrick, S., Gelatt, C.D., Vecchi, M.P.: {Optimization by Simulated
  Annealing}. Science  \textbf{220}(4598),  671--680 (1983).
  \doi{10.2307/1690046}

\bibitem{Kluger2003}
Kluger, Y.: {Spectral biclustering of microarray data: coclustering genes and
  conditions}. Genome Research  \textbf{13}(4),  703--716 (2003).
  \doi{10.1101/gr.648603}

\bibitem{Kong2019}
Kong, Z., Huang, Q., Li, X.: {Bi-Phase evolutionary biclustering algorithm with
  the NSGA-II algorithm}. In: IEEE International Conference on Advanced
  Robotics and Mechatronics (ICARM). pp. 146--149. IEEE (2019).
  \doi{10.1109/ICARM.2019.8834068}

\bibitem{Laguna2006}
Laguna, M., Mart{\'{i}}, R.: {Scatter Search}. In: Metaheuristic Procedures for
  Training Neutral Networks, pp. 139--152. Springer US (2006).
  \doi{10.1007/0-387-33416-5\_7}

\bibitem{Lashkargir2009}
Lashkargir, M., Monadjemi, S.A., Dastjerdi, A.B.: {A new biclustering method
  for gene expression data based on adaptive multiobjective particle swarm
  optimization}. In: International Conference on Computer and Electrical
  Engineering. pp. 559--563. IEEE (2009). \doi{10.1109/ICCEE.2009.245}

\bibitem{Li2009}
Li, G., Ma, Q., Tang, H., Paterson, A.H., Xu, Y.: {QUBIC: a qualitative
  biclustering algorithm for analyses of gene expression data}. Nucleic Acids
  Research  \textbf{37}(15),  e101--e101 (aug 2009). \doi{10.1093/nar/gkp491}

\bibitem{Liu2009}
Liu, J., Li, Z., Hu, X., Chen, Y.: {Biclustering of microarray data with MOSPO
  based on crowding distance}. BMC Bioinformatics  \textbf{10}(S4), ~S9 (2009).
  \doi{10.1186/1471-2105-10-S4-S9}

\bibitem{Liu2009b}
Liu, J., Li, Z., Hu, X., Chen, Y.: {Multi-objective ant colony optimization
  biclustering of microarray data}. In: IEEE International Conference on
  Granular Computing. pp. 424--429. IEEE (aug 2009).
  \doi{10.1109/GRC.2009.5255086}

\bibitem{Liu2008}
Liu, J., Li, Z., Liu, F., Chen, Y.: {Multi-objective particle swarm
  optimization biclustering of microarray data}. In: EEE International
  Conference on Bioinformatics and Biomedicine. pp. 363--366. IEEE (2008).
  \doi{10.1109/BIBM.2008.17}

\bibitem{Lu2015}
Lu, Y., Liu, Y.: {Biclustering of the gene expression data by coevolution
  cuckoo search}. International Journal Bioautomation  \textbf{19}(2),
  161--176 (2015)

\bibitem{Maatouk2014}
Ma{\^{a}}touk, O., Ayadi, W., Bouziri, H., Duval, B.: {Evolutionary algorithm
  based on new crossover for the biclustering of gene expression data}. In:
  International Conference on Pattern Recognition in Bioinformatics. pp. 48--59
  (2014). \doi{10.1007/978-3-319-09192-1\_5}

\bibitem{Maatouk2019}
Ma{\^{a}}touk, O., Ayadi, W., Bouziri, H., Duval, B.: {Evolutionary
  biclustering algorithms: an experimental study on microarray data}. Soft
  Computing  \textbf{23}(17),  7671--7697 (2019).
  \doi{10.1007/s00500-018-3394-4}

\bibitem{Maatouk2021}
Ma{\^{a}}touk, O., Ayadi, W., Bouziri, H., Duval, B.: {Evolutionary local
  search algorithm for the biclustering of gene expression data based on
  biological knowledge}. Applied Soft Computing  \textbf{104},  107177 (2021).
  \doi{10.1016/j.asoc.2021.107177}

\bibitem{madeira2004biclustering}
Madeira, S.C., Oliveira, A.L.: Biclustering algorithms for biological data
  analysis: a survey. IEEE/ACM transactions on computational biology and
  bioinformatics  \textbf{1}(1),  24--45 (2004)

\bibitem{Maulik2009}
Maulik, U., Mukhopadhyay, A., Bandyopadhyay, S.: {Finding multiple coherent
  biclusters in microarray data using variable string length multiobjective
  genetic algorithm}. IEEE Transactions on Information Technology in
  Biomedicine  \textbf{13}(6),  969--975 (2009).
  \doi{10.1109/TITB.2009.2017527}

\bibitem{Maulik2008}
Maulik, U., Mukhopadhyay, A., Bandyopadhyay, S., Zhang, M.Q., Zhang, X.:
  {Multiobjective fuzzy biclustering in microarray data: Method and a new
  performance measure}. In: IEEE Congress on Evolutionary Computation. pp.
  1536--1543. IEEE (2008). \doi{10.1109/CEC.2008.4630996}

\bibitem{Mitra2006}
Mitra, S., Banka, H.: {Multi-objective evolutionary biclustering of gene
  expression data}. Pattern Recognition  \textbf{39}(12),  2464--2477 (2006).
  \doi{10.1016/j.patcog.2006.03.003}

\bibitem{Mukhopadhyay2009}
Mukhopadhyay, A., Maulik, U., Bandyopadhyay, S.: {A novel coherence measure for
  discovering scaling biclusters from gene expression data}. Journal of
  Bioinformatics and Computational Biology  \textbf{07}(05),  853--868 (2009).
  \doi{10.1142/S0219720009004370}

\bibitem{Narmadha2021}
Narmadha, N., Rathipriya, R.: {Gene ontology analysis of gene expression data
  using hybridized PSO triclustering}. In: Machine Learning and Big Data
  Analytics Paradigms: Analysis, Applications and Challenges. pp. 437--466
  (2021). \doi{10.1007/978-3-030-59338-4\_22}

\bibitem{Nepomuceno2010}
Nepomuceno, J.A., Troncos, A., Aguilar-Ruiz, J.S.: {Evolutionary metaheuristic
  for biclustering based on linear correlations among genes}. In: ACM Symposium
  on Applied Computing. p.~1143. ACM Press (2010).
  \doi{10.1145/1774088.1774329}

\bibitem{Nepomuceno2009}
Nepomuceno, J.A., Troncoso, A., Aguilar-Ruiz, J.S.: {A hybrid metaheuristic for
  biclustering based on scatter search and genetic algorithms}. In:
  International Conference on Pattern Recognition in Bioinformatics. pp.
  199--210 (2009). \doi{10.1007/978-3-642-04031-3\_18}

\bibitem{Nepomuceno2011}
Nepomuceno, J.A., Troncoso, A., Aguilar-Ruiz, J.S.: {Biclustering of gene
  expression data by correlation-based scatter search}. BioData Mining
  \textbf{4}(1), ~3 (2011). \doi{10.1186/1756-0381-4-3}

\bibitem{Nepomuceno2015a}
Nepomuceno, J.A., Troncoso, A., Aguilar-Ruiz, J.S.: {Scatter search-based
  identification of local patterns with positive and negative correlations in
  gene expression data}. Applied Soft Computing  \textbf{35},  637--651 (2015).
  \doi{10.1016/j.asoc.2015.06.019}

\bibitem{Nepomuceno2015}
Nepomuceno, J.A., Troncoso, A., Nepomuceno-Chamorro, I.A., Aguilar-Ruiz, J.S.:
  {Integrating biological knowledge based on functional annotations for
  biclustering of gene expression data}. Computer Methods and Programs in
  Biomedicine  \textbf{119}(3),  163--180 (2015).
  \doi{10.1016/j.cmpb.2015.02.010}

\bibitem{Nepomuceno2018}
Nepomuceno, J.A., Troncoso, A., Nepomuceno-Chamorro, I.A., Aguilar-Ruiz, J.S.:
  {Pairwise gene GO-based measures for biclustering of high-dimensional
  expression data}. BioData Mining  \textbf{11}(1), ~4 (2018).
  \doi{10.1186/s13040-018-0165-9}

\bibitem{Pontes2013}
Pontes, B., Gir{\'{a}}ldez, R., Aguilar-Ruiz, J.S.: {Configurable pattern-based
  evolutionary biclustering of gene expression data}. Algorithms for Molecular
  Biology  \textbf{8}(1), ~4 (2013). \doi{10.1186/1748-7188-8-4}

\bibitem{Pontes2015}
Pontes, B., Gir{\'{a}}ldez, R., Aguilar-Ruiz, J.S.: {Biclustering on expression
  data: A review}. Journal of Biomedical Informatics  \textbf{57},  163--180
  (oct 2015). \doi{10.1016/j.jbi.2015.06.028}

\bibitem{pontes2015biclustering}
Pontes, B., Gir{\'a}ldez, R., Aguilar-Ruiz, J.S.: Biclustering on expression
  data: A review. Journal of biomedical informatics  \textbf{57},  163--180
  (2015)

\bibitem{Prelic2006}
Preli{\'{c}}, A., Bleuler, S., Zimmermann, P., Wille, A., B{\"{u}}hlmann, P.,
  Gruissem, W., Hennig, L., Thiele, L., Zitzler, E.: {A systematic comparison
  and evaluation of biclustering methods for gene expression data}.
  Bioinformatics  \textbf{22}(9),  1122--1129 (may 2006).
  \doi{10.1093/bioinformatics/btl060}

\bibitem{Zhang2007}
{Qingfu Zhang}, {Hui Li}: {MOEA/D: A multiobjective evolutionary algorithm
  based on decomposition}. IEEE Transactions on Evolutionary Computation
  \textbf{11}(6),  712--731 (dec 2007). \doi{10.1109/TEVC.2007.892759}

\bibitem{QinghuaHuang2012}
{Qinghua Huang}, {Dacheng Tao}, {Xuelong Li}, Liew, A.: {Parallelized
  evolutionary learning for detection of biclusters in gene expression data}.
  IEEE/ACM Transactions on Computational Biology and Bioinformatics
  \textbf{9}(2),  560--570 (2012). \doi{10.1109/TCBB.2011.53}

\bibitem{Rathipriya2011}
Rathipriya, R., Thangavel, K., Bagyamani, J.: {Binary particle swarm
  optimization based biclustering of web usage data}. International Journal of
  Computer Applications  \textbf{25}(2),  43--49 (2011).
  \doi{10.5120/3001-4036}

\bibitem{Reyes2006}
Reyes-Sierra, M., Coello, C.C., et~al.: Multi-objective particle swarm
  optimizers: A survey of the state-of-the-art. International journal of
  computational intelligence research  \textbf{2}(3),  287--308 (2006)

\bibitem{Rodriguez-Baena2011}
Rodriguez-Baena, D.S., Perez-Pulido, A.J., Aguilar-Ruiz, J.S.: {A biclustering
  algorithm for extracting bit-patterns from binary datasets}. Bioinformatics
  \textbf{27}(19),  2738--2745 (2011). \doi{10.1093/bioinformatics/btr464}

\bibitem{Sahoo2016}
Sahoo, P., Acharya, S., Saha, S.: {Automatic generation of biclusters from gene
  expression data using multi-objective simulated annealing approach}. In:
  International Conference on Pattern Recognition. pp. 2174--2179. IEEE (2016).
  \doi{10.1109/ICPR.2016.7899958}

\bibitem{Seridi2011}
Seridi, K., Jourdan, L., Talbi, E.G.: {Multi-objective evolutionary algorithm
  for biclustering in microarrays data}. In: IEEE Congress of Evolutionary
  Computation. pp. 2593--2599. IEEE (2011). \doi{10.1109/CEC.2011.5949941}

\bibitem{Seridi2015}
Seridi, K., Jourdan, L., Talbi, E.G.: {Using multiobjective optimization for
  biclustering microarray data}. Applied Soft Computing  \textbf{33},  239--249
  (2015). \doi{10.1016/j.asoc.2015.03.060}

\bibitem{Serin2011}
Serin, A., Vingron, M.: {DeBi: Discovering differentially expressed biclusters
  using a frequent itemset approach}. Algorithms for Molecular Biology
  \textbf{6}(1), ~18 (2011). \doi{10.1186/1748-7188-6-18}

\bibitem{Shabalin2009}
Shabalin, A.A., Weigman, V.J., Perou, C.M., Nobel, A.B.: {Finding large average
  submatrices in high dimensional data}. The Annals of Applied Statistics
  \textbf{3}(3) (2009). \doi{10.1214/09-AOAS239}

\bibitem{Talbi2009}
Talbi, E.G.: {Metaheuristics from design to implementation}. John Wiley and
  Sons (2009)

\bibitem{Tanay2002}
Tanay, A., Sharan, R., Shamir, R.: {Discovering statistically significant
  biclusters in gene expression data}. Bioinformatics  \textbf{18}(Suppl 1),
  S136--S144 (2002). \doi{10.1093/bioinformatics/18.suppl\_1.S136}

\bibitem{Teng2008}
Teng, L., Chan, L.: {Discovering biclusters by iteratively sorting with
  weighted correlation coefficient in gene expression data}. Journal of Signal
  Processing Systems  \textbf{50},  267--280 (2008).
  \doi{10.1007/s11265-007-0121-2}

\bibitem{Thangavel2012}
Thangavel, K., Bagyamani, J., Rathipriya, R.: {Novel hybrid PSO-SA model for
  biclustering of expression data}. Procedia Engineering  \textbf{30},
  1048--1055 (2012). \doi{10.1016/j.proeng.2012.01.962}

\bibitem{vandromme2020biclustering}
Vandromme, M., Jacques, J., Taillard, J., Jourdan, L., Dhaenens, C.: A
  biclustering method for heterogeneous and temporal medical data. IEEE
  Transactions on Knowledge and Data Engineering  (2020)

\bibitem{Xie2019}
Xie, J., Ma, A., Fennell, A., Ma, Q., Zhao, J.: {It is time to apply
  biclustering: a comprehensive review of biclustering applications in
  biological and biomedical data}. Briefings in Bioinformatics  \textbf{20}(4),
   1450--1465 (jul 2019). \doi{10.1093/bib/bby014}

\bibitem{Yang2014}
Yang, X.S., Deb, S.: {Cuckoo search: recent advances and applications}. Neural
  Computing and Applications  \textbf{24}(1),  169--174 (2014).
  \doi{10.1007/s00521-013-1367-1}

\bibitem{Yin2018}
Yin, L., Qiu, J., Gao, S.: {Biclustering of gene expression data using cuckoo
  search and genetic algorithm}. International Journal of Pattern Recognition
  and Artificial Intelligence  \textbf{32}(11),  1850039 (2018).
  \doi{10.1142/S0218001418500398}

\bibitem{Zitzler2004}
Zitzler, E., K{\"{u}}nzli, S.: {Indicator-based selection in multiobjective
  search}. In: Parallel Problem Solving from Nature - PPSN VIII. Lecture Notes
  in Computer Science, vol.~3242, pp. 832--842. Springer-Verlag (2004).
  \doi{10.1007/978-3-540-30217-9\_84}

\bibitem{Zitzler2001}
Zitzler, E., Laumanns, M., Thiele, L.: {SPEA2: Improving the strength Pareto
  evolutionary algorithm}. Tech. rep., Swiss Federal Institute Technology,
  Zurich, Switzerland (2001)

\end{thebibliography}

\end{document}